\documentclass[12pt]{iopart}

\usepackage{amssymb}
\usepackage[english]{babel}
\usepackage{bm}
\usepackage{booktabs}
\usepackage{graphicx}
\usepackage{hyperref}
\usepackage{xcolor}

\newcommand{\text}[1]{\mathrm{#1}}

\newcommand{\dd}{\mathop{}\!\mathrm{d}}
\newcommand{\E}{\mathbb{E}}
\newcommand{\Ls}{\mathcal{L}}
\newcommand{\pp}{\mathop{}\!{\partial}}
\newcommand{\vx}{{\bm{x}}}
\newcommand{\vy}{{\bm{y}}}
\newcommand{\vz}{{\bm{z}}}

\begin{document}

\title[RG-Flow: A hierarchical and explainable flow model]{RG-Flow: A hierarchical and explainable flow model based on renormalization group and sparse prior}

\author{Hong-Ye Hu\textsuperscript{1}, Dian Wu\textsuperscript{2}, Yi-Zhuang You\textsuperscript{3}, Bruno Olshausen\textsuperscript{4}, and Yubei Chen\textsuperscript{5}}

\address{\textsuperscript{1} \textsuperscript{3} Department of Physics, University of California, San Diego, La Jolla, \\ CA~92093}
\address{\textsuperscript{1} Department of Physics, Harvard University, 17 Oxford Street, Cambridge, MA~02138}
\address{\textsuperscript{2} Computational Quantum Science Laboratory, École polytechnique fédérale de Lausanne (EPFL) Station 3, CH-1015 Lausanne}
\address{\textsuperscript{4} \textsuperscript{5} Redwood Center, Berkeley AI Research, University of California, Berkeley, Berkeley, CA~94720}
\ead{\mailto{hyhu@ucsd.edu}, \mailto{dian.wu@epfl.ch}, \mailto{yzyou@ucsd.edu}, \mailto{baolshausen@berkeley.edu}, \mailto{yubeic@berkeley.edu}}

\begin{abstract}
Flow-based generative models have become an important class of unsupervised learning approaches. In this work, we incorporate the key ideas of renormalization group (RG) and sparse prior distribution to design a hierarchical flow-based generative model, RG-Flow, which can separate information at different scales of images and extract disentangled representations at each scale. We demonstrate our method on synthetic multi-scale image datasets and the CelebA dataset, showing that the disentangled representations enable semantic manipulation and style mixing of the images at different scales. To visualize the latent representations, we introduce receptive fields for flow-based models and show that the receptive fields of RG-Flow are similar to those of convolutional neural networks. In addition, we replace the widely adopted isotropic Gaussian prior distribution by the sparse Laplacian distribution to further enhance the disentanglement of representations. From a theoretical perspective, our proposed method has $O(\log L)$ complexity for inpainting of an image with edge length $L$, compared to previous generative models with $O(L^2)$ complexity.

\end{abstract}

\section{Introduction}

One of the most important unsupervised learning tasks is to learn the data distribution and build generative models. Over the past few years, various types of generative models have been proposed. Flow-based generative models are a particular family of generative models with tractable distributions~\cite{DBLP:conf/iclr/DinhSB17, DBLP:conf/nips/KingmaD18, DBLP:conf/nips/ChenRBD18, DBLP:conf/nips/ChenBDJ19, DBLP:conf/icml/BehrmannGCDJ19, DBLP:conf/icml/HoogeboomBW19, DBLP:journals/corr/abs-2003-13913, DBLP:journals/corr/abs-2002-02428, DBLP:conf/nips/KaramiSSDD19}, yet their latent variables are on equal footing and mixed globally. Here, we propose a new flow-based model, RG-Flow, which is inspired by the idea of \emph{renormalization group} in statistical physics. RG-Flow imposes locality and hierarchical architecture in bijective transformations. It allows us to access and modify information in input data at different scales by controlling different latent variables, which offers better explainability. Combined with \emph{sparse prior distributions}~\cite{olshausen1996emergence, olshausen1997sparse, hyvarinen2000independent}, we show that RG-Flow achieves hierarchical disentanglement of representations.

Renormalization group (RG) is a powerful tool in physics to analyze statistical mechanics systems and quantum field theories~\cite{PhysicsPhysiqueFizika.2.263, PhysRevB.4.3174}. It progressively extracts more coarse-scale statistical features of the physical system and decimates irrelevant fine-grained statistics at each scale. Typically, the transformations in RG are local, not bijective, and designed by physicists. Flow-based models also use cascaded transformations to progressively turn a complicated data distribution into a simple distribution, such as a Gaussian, but on the contrary, those transformations are usually global, bijective, and automatically learned. In this work, we propose RG-Flow, which combines the key ideas from both RG and flow-based models. RG-Flow learns the optimal RG transformation from data using local bijective transformations, and also serves as a hierarchical generative model for the data. Latent representations are extracted gradually at different scales to capture the statistical features of the data at the corresponding scales, and supplemented jointly to invert the transformations when generate the data. This method has been recently explored in the physics community as NeuralRG~\cite{PhysRevLett.121.260601, PhysRevResearch.2.023369}.

Our main contributions are two-fold: First, we point out that RG-Flow can naturally separate the information at different scales in the input image distribution, and assign hierarchical latent variables on a hyperbolic tree. Taking the human face dataset CelebA~\cite{DBLP:conf/iccv/LiuLWT15} as an example, the network will not only find high-level representations like the gender and the emotion, but also mid-level and low-level ones like the shape of a single eye. To visualize representations at different scales, we adopt the concept of \emph{receptive field} from convolutional neural networks (CNN)~\cite{LeCun1988ATF, DBLP:conf/nips/CunBDHHHJ89} and visualize their structures in RG-Flow. In addition, since the representations are separated in a hierarchical fashion, they can be mixed to different extents at different scales, thus separate the mixing of content and style in images. Second, we find that the sparse prior distribution is helpful to further disentangle representations and make them more explainable. As the widely adopted Gaussian prior is rotationally symmetric, each latent variable in a flow model can be arbitrarily mixed with others and lose any clear semantic meaning. Using a sparse prior, we clearly demonstrate the semantic meaning of the latent space.

\section{Related work}

Some flow-based generative models also possess multi-scale latent space~\cite{DBLP:conf/iclr/DinhSB17, DBLP:conf/nips/KingmaD18, yu2020wavelet, bhattacharyya2020normalizing, camuto2021learning, voleti2021multi}, and recently hierarchies of features have been utilized in~\cite{DBLP:journals/corr/abs-2006-10848}, where the high-level features are shown to perform strongly in out-of-distribution detection task. However, previous models do not impose hard locality constraint in their multi-scale architectures. In~\ref{append:RNVP}, the differences between globally connected multi-scale flows and RG-Flow are discussed, and we find that semantically meaningful receptive fields do not show up in the former. Recently, other more expressive bijective maps have been developed~\cite{DBLP:conf/icml/HoogeboomBW19, DBLP:conf/nips/KaramiSSDD19, DBLP:conf/nips/DurkanB0P19}, and those methods can be incorporated into our proposed architecture to further improve the expressive power of RG-Flow.

Some other classes of generative models rely on a separate inference model to obtain the latent representation. Examples include variational autoencoders~\cite{DBLP:journals/corr/KingmaW13}, adversarial autoencoders~\cite{DBLP:journals/corr/MakhzaniSJG15}, InfoGAN~\cite{DBLP:conf/nips/ChenCDHSSA16}, BiGAN~\cite{DBLP:conf/iclr/DonahueKD17, DBLP:conf/iclr/DumoulinBPLAMC17}, and progressive growing GAN~\cite{karras2017progressive}. Those techniques typically do not arrange latent variables in a hierarchical way, and their inference of latent variables is approximate. Notably, recent advances have suggested that introducing hierarchical latent variables in them can be beneficial~\cite{DBLP:journals/corr/abs-2007-03898}. In addition, the coarse-to-fine fashion of the generation process has also been discussed in other generative models, such as Laplacian pyramid of adversarial networks~\cite{DBLP:conf/nips/DentonCSF15}, and multi-scale autoregressive models~\cite{DBLP:conf/icml/ReedOKCWCBF17}.

\emph{Disentangled representations}~\cite{DBLP:journals/neco/TenenbaumF00, DiCarlo_2007, Bengio_2013, DBLP:conf/pkdd/MolnarCB20} are another important aspect in understanding how a model generates images~\cite{DBLP:journals/corr/abs-1812-02230}. Especially, disentangled high-level representations have been discussed and improved from information theoretical principles~\cite{DBLP:journals/corr/CheungLBO14, DBLP:conf/nips/ChenCDHSSA16, DBLP:conf/nips/ChenLGD18, DBLP:conf/iclr/HigginsMPBGBML17, DBLP:conf/iclr/KipfPW20, DBLP:conf/icml/KimM18, DBLP:conf/icml/LocatelloBLRGSB19, DBLP:journals/corr/abs-1812-01161}. Moreover, the multi-scale structure appears intrinsically in the distribution of natural images. Apart from generating whole images, the multi-scale representations can be used to perform other tasks, such as style transfer~\cite{7780634, CycleGAN2017}, content mixing~\cite{DBLP:conf/cvpr/KarrasLA19, DBLP:journals/corr/abs-1911-13270, DBLP:conf/cvpr/KarrasLAHLA20}, and texture synthesis~\cite{DBLP:conf/icml/BergmannJV17, DBLP:journals/corr/JetchevBV16, DBLP:conf/nips/GatysEB15, DBLP:conf/eccv/JohnsonAF16, DBLP:conf/icml/UlyanovLVL16}.

Typically, in flow-based generative models, Gaussian distribution is used as the prior distribution for the latent space. Due to the rotational symmetry of Gaussian prior, an arbitrary rotation of the latent space leaves the likelihood unchanged, therefore the latent variables can be arbitrarily mixed during the training and lose their semantic meanings. Sparse priors~\cite{olshausen1996emergence, olshausen1997sparse, hyvarinen2000independent} have been proposed as an important tool for unsupervised learning, which leads to better explainability in various domains~\cite{ainsworth2018oi, arora2018linear, DBLP:journals/corr/abs-1910-03833}. To break the symmetry of Gaussian prior and improve the explainability, we incorporate the sparse Laplacian prior in RG-Flow. Please refer to \fref{fig:toy_model} for a quick illustration on the difference between Gaussian prior and the sparse prior, where the sparse prior intuitively leads to better disentanglement of the features.

\emph{Renormalization group} (RG) has a broad impact ranging from particle physics to astrophysics. Apart from the analytical studies in field theories~\cite{PhysRevB.4.3174, RevModPhys.70.653, RevModPhys.71.S358}, RG has also been useful in numerically simulating quantum states. The multi-scale entanglement renormalization ansatz (MERA)~\cite{PhysRevLett.101.110501, PhysRevLett.112.240502} implements the hierarchical architecture of RG in tensor networks to represent quantum states. The exact holographic mapping (EHM)~\cite{Qi2013ExactHM, PhysRevB.93.035112, PhysRevB.93.104205} further extends MERA to a bijective (unitary) flow model between the latent product state and the observable entangled state. Recently, the MERA architecture and deep neural networks have been incorporated to design a flow-base generative model that learns EHM from statistical physics and quantum field theory actions~\cite{PhysRevLett.121.260601, PhysRevResearch.2.023369}. In quantum machine learning, recent development of quantum convolutional neural networks have also utilized the MERA architecture~\cite{Cong2019}. The similarity between RG and deep learning has been discussed in several works~\cite{beny2013deep, mehta2014exact, Beny2015, oprisa2017criticality, Lin2017, Gan_2017}, and information theoretic objectives to guide machine-learning RG transformations have been proposed in recent works~\cite{Koch-Janusz2018, PhysRevResearch.2.023369, PhysRevX.10.011037}. The semantic meaning of the emergent latent space has been related to quantum gravity~\cite{PhysRevD.86.065007, Pastawski2015}, which leads to the exciting development of machine-learning holography~\cite{PhysRevB.97.045153, PhysRevD.98.106014, PhysRevD.99.106017, Akutagawa_2020, hashimoto2020neural}.

\section{Methods}

\subsection{Flow-based generative models}

\begin{figure}[htb]
\centering
\includegraphics[width=0.7\linewidth]{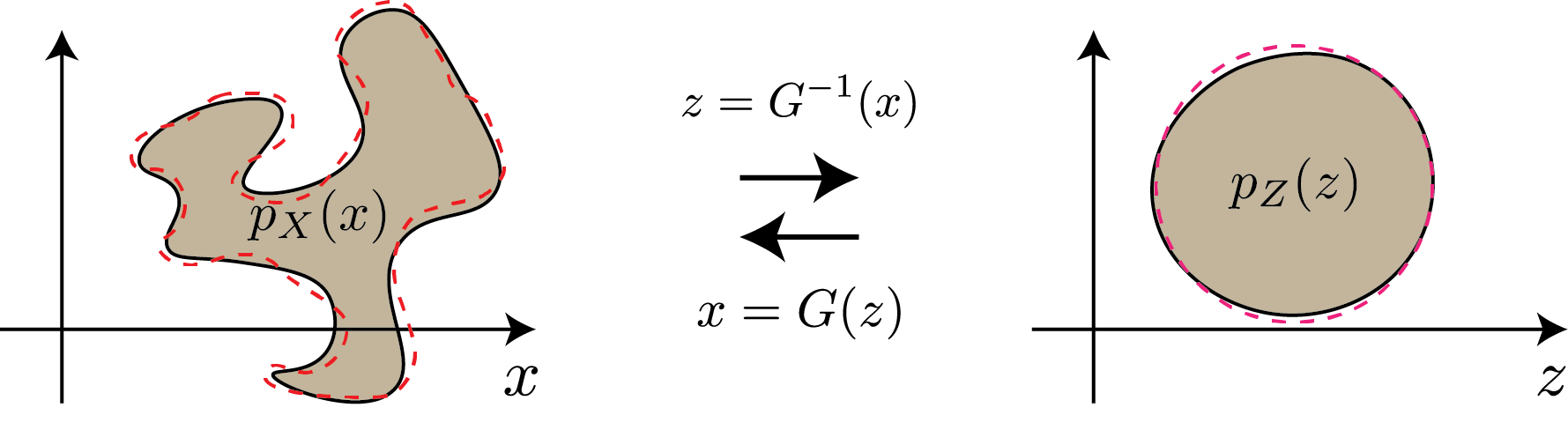}
\caption{The observable variables $\vx$ are the data that we want to model, which may follow a complicated probability distribution $p_X(\vx)$. The goal of flow-based models is to learn a bijective mapping $G$ between the observable space and the latent representation such that the latent variables follow a simple distribution $p_Z(\vz)$. The red dash circle on the right shows the ideal target distribution $p_Z^{*}(\vz)$ in the latent space. The red dash circle on the left shows the inverse target distribution $p_X^{*}(\vx)$ in the data space. The loss function measures the discrepancy between the data distribution $p_X(\vx)$ and the target distribution $p_X^{*}(\vx)$. }
\label{fig:flow}
\end{figure}

Flow-based generative models are a family of generative models with tractable distributions, which allows exact and efficient sampling and evaluation of the probability density~\cite{DBLP:journals/corr/DinhKB14, DBLP:conf/iclr/DinhSB17, DBLP:conf/nips/KingmaD18, DBLP:conf/nips/ChenBDJ19}. The key idea is to build a bijective map $\vx = G(\vz)$ between the observable variables $\vx$ and the latent variables $\vz$, where $\vz$ has a simple distribution $p_Z(\vz)$, e.g. Gaussian, as shown in \fref{fig:flow}. The latent variables $\vz$ usually have a simple distribution that can be easily sampled, for example the i.i.d. Gaussian distribution. In this way, the data can be efficiently generated by first sampling $\vz$ and then mapping them to $\vx$. Due to the equality $p_X(\vx) \dd \vx = p_Z(\vz) \dd \vz \Rightarrow p_X(\vx) = p_Z(\vz) \left| \frac{\pp \vz}{\pp \vx} \right|$, the log-probability density $\log p_X(\vx)$ is given by
\begin{eqnarray}
\log p_X(\vx) &= \log p_Z(\vz) + \log \left| \det \frac{\pp G^{-1}(\vx)}{\pp \vx} \right| \\
&= \log p_Z(\vz) - \log \left| \det \frac{\pp G(\vz)}{\pp \vz} \right|.
\label{eq:px_pz}
\end{eqnarray}
The bijective map is usually composed from a series of bijectors, $G(\vz) = G_1 \circ G_2 \circ \cdots \circ G_n(\vz)$, where each bijector $G_i$ has a tractable Jacobian determinant and its inverse $R_i = G_i^{-1}$ can be computed efficiently. $R$ stands for ``representation'' and $G$ stands for ``generation''. The two key ingredients in flow-based models are the design of the bijective map $G$ and the choice of the prior distribution $p_Z$.

\subsection{Structure of RG-Flow}

Much of previous research has focused on designing more powerful bijective blocks for the generator $G$ to improve its expressive power and achieve better approximations of complicated probability distributions. In this work, we focus on designing an architecture that arranges the bijective blocks in a hierarchical fashion to separate features at different scales in the data and disentangle latent representations.

\begin{figure}[htb]
\centering
\includegraphics[width=\linewidth]{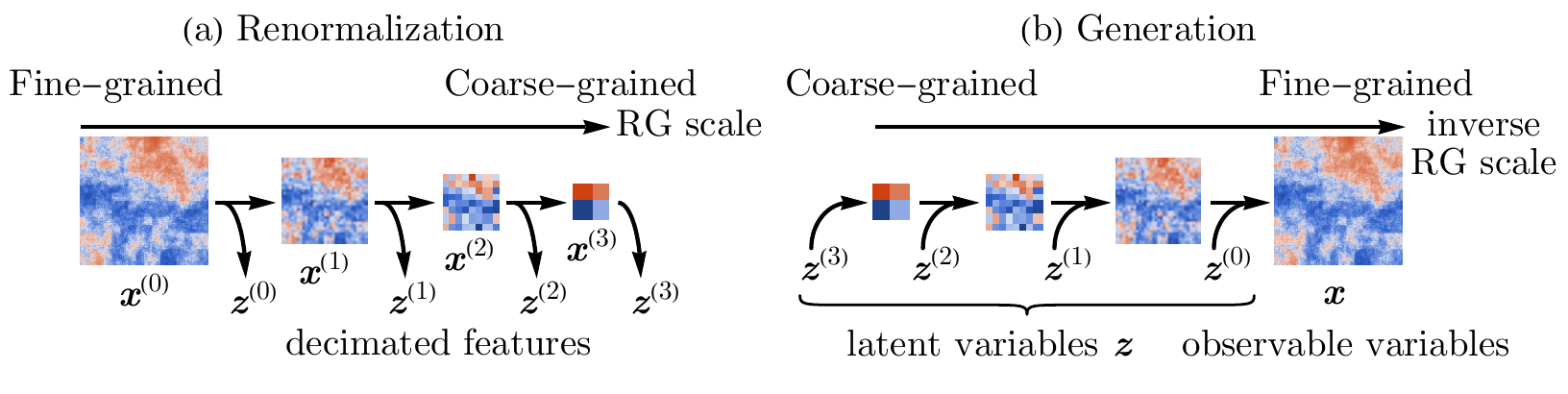}
\caption{
(a) Each layer of forward RG transformation extracts coarse-grained information (decimation) to send to the next level for further processing and splits out the independent fine-grained features, which is to be left at the current stage. RG progressively builds a hierarchical representation from the input image at different scales.
(b) The inverse RG transformation generates the fine-grained image from latent variables. At each layer, the coarse-grained information from higher level is merged with the independent fine-grained information at the current level, and then the inverse transform of the merged representation is sent to the lower level.
}
\label{fig:idea_of_RG}
\end{figure}

Our design is motivated by the idea of RG in physics, which progressively separates the coarse-grained information in input data from independent fine-grained information by local transformations at different scales. Let $\vx$ be the observable variables, or the input image (level $0$), denoted as $\vx^{(0)} \equiv \vx$. A step of the RG transformation extracts the coarse-grained information $\vx^{(1)}$, send it to the next layer (level $1$), and splits out the remaining independent fine-grained information as auxiliary variables $\vz^{(0)}$, which is left to the current level. This procedure is described by the following recurrence equation (at level $h$ for example):
\begin{equation}
\vx^{(h + 1)}, \vz^{(h)} = R_h(\vx^{(h)}),
\end{equation}
which is illustrated in \fref{fig:idea_of_RG}~(a), where $\dim(\vx^{(h + 1)}) + \dim(\vz^{(h)}) = \dim(\vx^{(h)})$. At each level, the transformation $R_h$ is a local bijective map, which is constructed by stacking trainable bijective blocks, and we will specify its details later. The split-out information $\vz^{(h)}$ can be viewed as latent variables arranged at the corresponding scale. Following the bijectivity of $R_h$, the inverse RG transformation step $G_h \equiv R_h^{-1}$ generates the fine-grained image:
\begin{equation}
\vx^{(h)} = R_h^{-1}(\vx^{(h + 1)}, \vz^{(h)}) = G_h(\vx^{(h + 1)}, \vz^{(h)}).
\end{equation}
At the highest level $h_L$, the most coarse-grained image $\vx^{(h_L)} = G_{h_L}(\vz^{(h_L)})$ can be considered as generated directly from the latent variables $\vz^{(h_L)}$ without referring to any higher-level image, where $h_L = \log_2 L - \log_2 m$ for the original image of size $L \times L$ and the local transformations acting on kernel size $m \times m$. Therefore, given the latent variables $\vz = \{\vz^{(h)}\}$ at all levels $h$, the original image can be restored by the following composite map:
\begin{equation}
\vx \equiv \vx^{(0)} = G_0(G_1(G_2(\ldots, \vz^{(2)}), \vz^{(1)}), \vz^{(0)}) \equiv G(\vz),
\end{equation}
as illustrated in \fref{fig:idea_of_RG}~(b). RG-Flow is a flow-based generative model that uses the above composite bijective map $G$ as the generator.

\begin{figure}[htb]
\centering
\includegraphics[width=\linewidth]{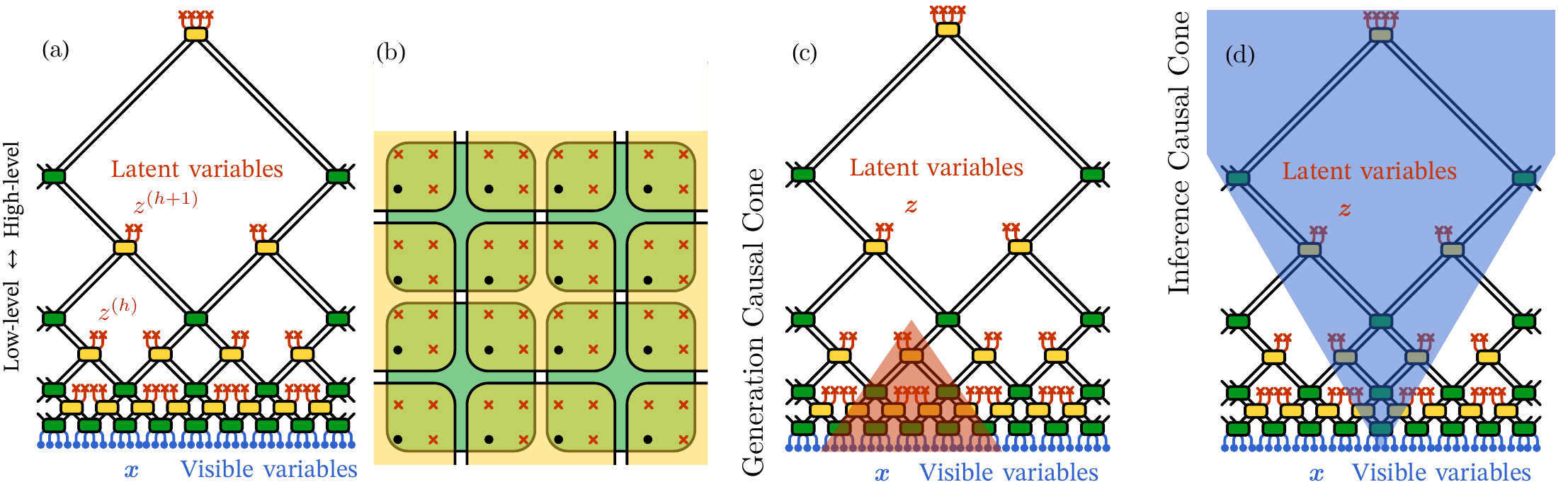}
\caption{
(a) Side view of our network. The green and the yellow blocks denote the disentanglers and the decimators respectively, which are bijective maps.
(b) Top-down view of our network, which shows the overlaps between the disentanglers and the decimators in two adjacent layers.
(c) The red area shows the generation causal cone for a latent variable.
(d) The blue area shows the inference causal cone for an observable variable.
}
\label{fig:structure}
\end{figure}

To model the RG transformation, we arrange the bijective blocks in a hierarchical network architecture. \fref{fig:structure}~(a) shows the side view of the network, where each green or yellow block is a local bijective map. Following the notation of MERA, the green blocks are the \emph{disentanglers}, which reparametrize local variables to reduce their correlations, and the yellow blocks are the \emph{decimators}, which split the decimated features out as latent variables. If there are no disentanglers, the structure would simply be a tree. Disentanglers are added to reduce the local correlations of variables and encourage the decimated variables to be noise-like fine-grained information. The blue dots on the bottom are the observable variables $\vx$, or the input image, and the red crosses are the latent variables $\vz$. We omit color channels of the images at each level in the illustration, since we keep the number of channels unchanged through the transformation.

\Fref{fig:structure}~(b) shows the top-down view of a step of the RG transformation. The disentanglers (green blocks) and the decimators (yellow blocks) are interwoven on top of each other. The covered area of a disentangler or decimator is defined as the kernel size $m \times m$ of the bijector. For example, in \fref{fig:structure}~(b), the kernel size is $4 \times 4$. After each decimator, three fourth of the degrees of freedom are decimated into latent variables (red crosses), so the edge length of the image is halved.

As a precise mathematical description of \fref{fig:structure}, for a single RG transformation step $R_h$ including a disentangler layer and a decimator layer at level $h$, in a spatial block $(p, q)$ labeled by $p, q = 0, 1, \ldots, \frac{L}{2^h m} - 1$, the mapping from $\vx^{(h)}$ to ($\vx^{(h + 1)}, \vz^{(h)}$) is given by:
\begin{eqnarray}
&\left\{ \vy^{(h)}_{\mathbf{r}_{a, b}} \right\}_{(a, b) \in \square^1_m}
= R_h^\text{dis}\left( \left\{ \vx^{(h)}_{\mathbf{r}_{a, b}} \right\}_{(a, b) \in \square^1_m} \right), \\
&\left\{ \vx^{(h + 1)}_{\mathbf{r}'_{a, b}} \right\}_{(a, b) \in \square^2_m}, \left\{ \vz^{(h)}_{\mathbf{r}'_{a, b}} \right\}_{(a, b) \in \square^1_m / \square^2_m}
= R_h^\text{dec}\left( \left\{ \vy^{(h)}_{\mathbf{r}'_{a, b}} \right\}_{(a, b) \in \square^1_m} \right),
\end{eqnarray}
where $\vy$ is the intermediate result after the disentangler but not the decimator,
\begin{eqnarray}
\mathbf{r}_{a, b} &= 2^h (m p + \frac{m}{2} + a, m q + \frac{m}{2} + b), \\
\mathbf{r}'_{a, b} &= 2^h (m p + a, m q + b)
\end{eqnarray}
are pixel positions that the disentangler and the decimator act on respectively, and $\square^k_m = \{(k a, k b) \mid a, b = 0, 1, \ldots, \frac{m}{k} - 1\}$ denotes the set of pixels in a $m \times m$ square with stride $k$. The notation $\vx^{(h)}_{(i, j)}$ stands for the variable (a vector of all color channels) at the RG level $h$ and the spatial position $(i, j)$, and similarly for $\vy$ and $\vz$.

The disentanglers $R_h^\text{dis}$ and the decimators $R_h^\text{dec}$ can be any bijective neural network. Practically, We use the coupling layer proposed in the Real NVP architecture~\cite{DBLP:conf/iclr/DinhSB17} to build them, with a detailed description in~\ref{append:net}. By specifying the above RG transformation step $R_h = R_h^\text{dec} \circ R_h^\text{dis}$, the generator $G_h \equiv R_h^{-1}$ is automatically specified as the inverse transformation step.

\subsection{Training objective} \label{sec:loss}

After decomposing the input data into multiple scales, we still need to enforce that the latent variables at the same scale are disentangled. We take the usual approach that the latent variables $\vz$ are independent random variables, described by a factorized prior distribution
\begin{equation}
p_Z^{*}(\vz) = \prod_l p(z_l),
\end{equation}
where the index $l = (h, i, j, c)$ labels every latent variable by its RG level $h$, spatial position $(i, j)$, and color channel $c$. This prior gives the network the incentive to minimize the mutual information between latent variables. This \emph{minimal bulk mutual information} (minBMI) principle has been proposed to be the information theoretic principle that defines the RG transformation~\cite{PhysRevLett.121.260601, PhysRevResearch.2.023369}. The particular choice of the distribution $p(z_l)$ is discussed in~\ref{sec:sparse_prior}.

To model the data distribution, we minimize the negative log-likelihood of inputs $\vx$ sampled from the dataset. The loss function reads:
\begin{eqnarray}
\Ls &= D_\text{KL}\left( p_X(\vx) \mid p_X^{*}(\vx) \right) - \E_{\vx \sim p_X(\vx)} \log p_X(\vx) \label{eqn:DKL} \\
&= -\E_{\vx \sim p_X(\vx)} \log p_X^{*}(\vx) \\
&= -\E_{\vx \sim p_X(\vx)} \left( \log p_Z^{*}\left( R(\vx) \right) + \log \left| \det \frac{\pp R(\vx)}{\pp \vx} \right| \right),
\end{eqnarray}
where $R \equiv G^{-1}$ is the forward RG transformation and contains trainable parameters, and the representation $R(\vx) = \vz$ is the latent variables obtained by transforming the input sample. Please note that the second term of Equation~\ref{eqn:DKL} is a constant as it only involves the data distribution $p_X(\vx)$. The minimization of negative log-likelihood evaluated on training data is equivalent to the maximization of probability of generating those data from the neural network. As shown in Figure~\ref{fig:flow}, our goal is to transform a complicated data distribution $p_X(\vx)$ to a simple target distribution $p_Z^{*}(\vz)$ in the latent space, shown by the red dash circle on the right. The transformation $G$ inverts $p_Z^{*}(\vz)$ to $p_X^{*}(\vx)$ in the data space, which is shown by the red dash circle on the left. The maximum likelihood loss function measures the discrepancy between the target distribution $p_X^{*}(\vx)$ and the data distribution $p_X(\vx)$ by KL-divergence.

\subsection{Receptive fields of latent variables}

Considering the hierarchical and local architecture in our network, we define the \emph{generation causal cone} for a latent variable to be the affected area in the observable image when that latent variable is changed. This is illustrated as the red cone in \fref{fig:structure}~(c).

To visualize the representations in the latent space, we define the \emph{receptive field} for a latent variable $z_l$ as:
\begin{equation}
\text{RF}_l = \E_{\vz \sim p_Z(\vz)} \left| \frac{\pp G(\vz)}{\pp z_l} \right|_c,
\end{equation}
where $|\cdot|_c$ denotes the $1$-norm over the color channel dimension, so that $\text{RF}_l$ is a two-dimensional array for each $l$. The receptive field shows the response of the generated image $\vx = G(\vz)$ to an infinitesimal change of the latent variable $z_l$, averaged over the latent distribution $p_Z$. Following the definition, the receptive field of any latent variable is always contained in its generation causal cone, and the receptive fields of higher-level latent variables are larger than those of lower-level ones. In particular, if the receptive fields of two latent variables do not overlap, which is often the case at low levels, those latent variables automatically become disentangled.

\subsection{Image inpainting and error correction}

Another advantage of the locality in our network can be demonstrated in the inpainting task. Similar to the generation causal cone, we can define the \emph{inference causal cone} shown as the blue cone in \fref{fig:structure}~(d). If we perturb a pixel at the bottom of the inference causal cone, all the latent variables inside the cone will be affected, while the ones outside cannot be affected. An important property of the hyperbolic tree-like network is that higher levels contain exponentially fewer latent variables. Even though the inference causal cone expands as we reach higher levels, the number of latent variables dilutes exponentially as well, resulting in a constant number of latent variables covered by the inference causal cone at each level.

Therefore, if a small local region on the input image is corrupted, only $O(\log L)$ latent variables  can be affected by the corruption, where $L$ is the edge length of the entire image, and we only need to adjust those latent variables to inpaint the region. While for globally connected networks, all $O(L^2)$ latent variables have to be adjusted.

\subsection{Sparse prior distribution} \label{sec:sparse_prior}

In~\ref{sec:loss}, we have implemented the minBMI principle of RG transformation by using a factorized prior distribution, i.e. $p_Z^{*}(\vz) = \prod_l p(z_l)$. The common choice for $p(z_l)$ is the standard Gaussian distribution, which makes $p_Z^{*}(\vz)$ spherical symmetric. Therefore, if we apply an arbitrary rotation on the latent space, which transforms a latent vector $\vz$ to $\vz'$, the prior probability $p_Z^{*}(\vz')$ will remain the same as $p_Z^{*}(\vz)$, so there is no incentive for the network $R$ to map the input $\vx$ to $\vz$ rather than $\vz'$. However, each basis vector of $\vz'$ is a linear combination of those of $\vz$, so the latent representations can be arbitrarily mixed by the rotation.

To overcome this issue, we propose an anisotropic sparse prior distribution for $p_Z^{*}(\vz)$, which breaks the spherical symmetry of the latent space. The sparsity means that it has a heavy tail along each basis vector, incentivizing the network to map each semantically meaningful feature in the observable space to a single variable in the latent space, which helps the disentanglement of representations. In this work, we choose Laplacian distribution $p(z_l) = \frac{1}{2 b} \exp(-|z_l| / b)$, which is sparser than Gaussian distribution, as the former has a larger kurtosis than the latter. In~\ref{append:sparse_prior}, we show an example of a two-dimensional pinwheel distribution to illustrate this intuition.

\section{Experiments}

\begin{figure}[htb]
\centering
\includegraphics[width=\linewidth]{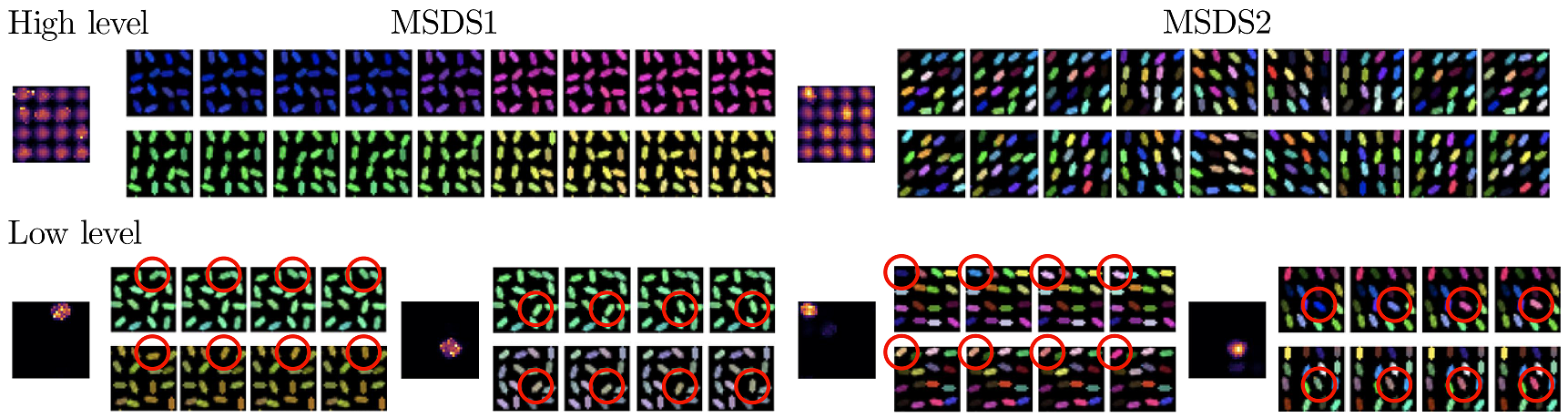}
\caption{Multi-scale latent representations of MSDS1 and MSDS2 learned by RG-Flow. In each group of subfigures, the one on the left shows the receptive field of a latent variable, and the two rows show the variation of two sample images when we vary that latent variable.}
\label{fig:msds_latent}
\end{figure}

\subsection{Synthetic multi-scale image datasets}

To illustrate RG-Flow's ability to disentangle representations at different scales and spatial positions, we propose two synthetic image datasets with multi-scale features, which we name MSDS1 and MSDS2. Their samples are shown in~\ref{append:MSDS}. In each image, there are $16$ ovals with different colors and orientations. In MSDS1, all ovals in a same image have almost the same color, while their orientations are randomly distributed, so the color is a global feature and the orientations are local ones. In MSDS2, on the contrary, the orientation is a global feature and the colors are local ones.

We implement RG-Flow as shown in \fref{fig:structure}. After training, we find that RG-Flow can easily capture the characteristics of those datasets. Namely, the ovals in each image from MSDS1 have almost the same color, and from MSDS2 the same orientation. In \fref{fig:msds_latent}, we plot the receptive fields of some latent variables at different levels, and the effect of varying them. For MSDS1, if we vary a high-level latent variable, the color of the whole image will change, which shows that the network has captured the global feature of this dataset. Meanwhile, if we vary a low-level latent variable, the orientation of only one oval at the corresponding position will change. As the ovals are spatially separated, the low-level representations of them are disentangled by the generation causal cones of RG-Flow. Similarly, for MSDS2, if we vary a high-level latent variable, the orientations of all ovals will change. If we vary a low-level latent variable, the color of only one oval will change.
For comparison, we also train Real NVP on the two datasets, and find that it fails to learn the global and the local characteristics of those datasets. Quantitative results are reported in~\ref{append:MSDS}.

\begin{figure}[htb]
\centering
\includegraphics[width=\linewidth]{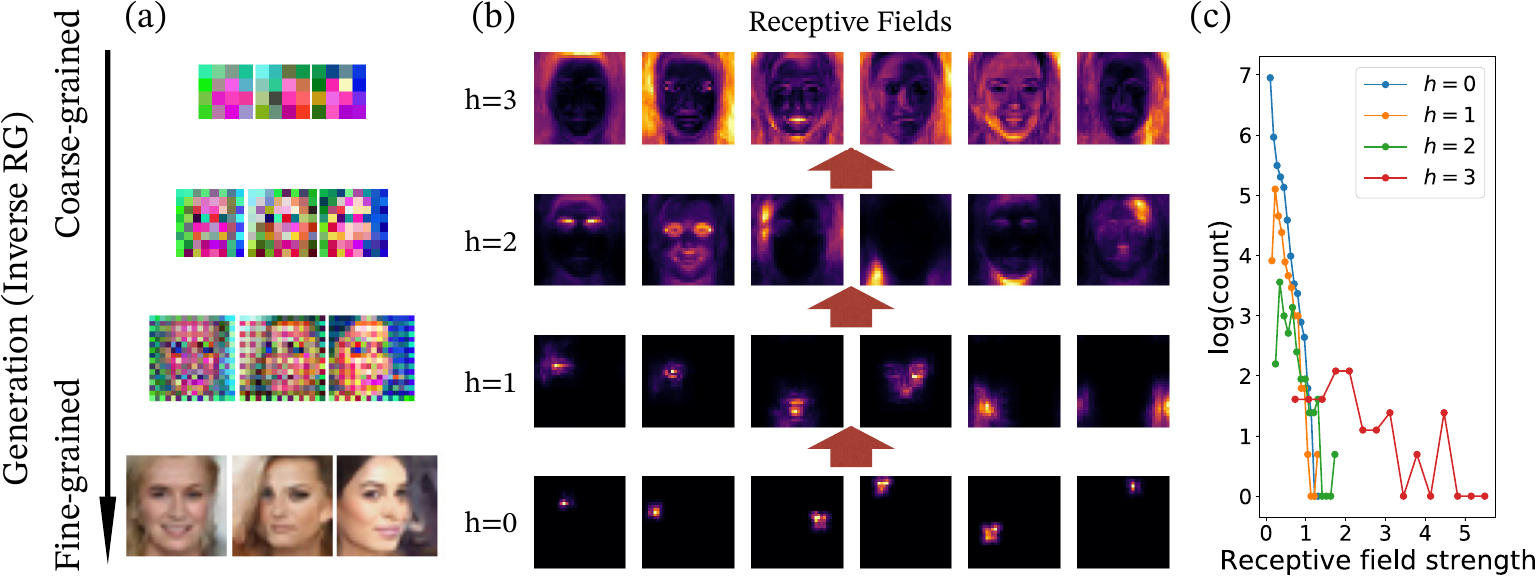}
\caption{
(a) Progressive generation of three sample images from CelebA using the inverse RG transformation. At high levels, the images being generated are coarse-grained. The inverse RG adds details to the images in each level, and the images become fine-grained at low levels.
(b) Receptive fields of typical latent variables from low levels to high levels. The strength of each receptive field is rescaled to one for better visualization.
(c) Histograms of receptive field strengths of latent variables at each level. The $x$-axis shows the receptive field strength, defined as the pixel-wise average of the latent variable's receptive field, and $y$-axis shows the logarithm of the number of latent variables in the histogram bin.
}
\label{fig:RF}
\end{figure}

\subsection{Human face dataset}

Next, we apply RG-Flow to more complicated multi-scale datasets. Most of our experiments use the human face dataset CelebA~\cite{DBLP:conf/iccv/LiuLWT15}, and we crop and scale the images to $32 \times 32$ pixels. Details of the network and the training procedure are described in~\ref{append:net}. Experiments on other datasets and quantitative evaluations are reported in~\ref{append:other_datasets}.

After training, the network learns to progressively generate finer-grained images, as shown in \fref{fig:RF}~(a). Note that the colors in the coarse-grained images are not necessarily the same as those at the corresponding positions in the fine-grained images, because there is no constraint to prevent the RG transformation from mixing color channels.

\subsection{Receptive fields}

To visualize the latent space representation, we compute the receptive field for each latent variable, and list some of them in \fref{fig:RF}~(b). The receptive size is small for low-level variables and large for high-level ones, as indicated from the generation causal cone. At the lowest level ($h = 0$), the receptive fields are merely small dots. At the second lowest level ($h = 1$), small structures emerge, such as an eyebrow, an eye, and a mouth. At the middle level ($h = 2$), there are more complex structures like a pair of eyebrows and a facial expression. At the highest level ($h = 3$), each receptive field grows to the whole image. We will investigate those explainable latent representations in~\ref{sec:feature}. For comparison, we show receptive fields of Real NVP in~\ref{append:RNVP}. Even though Real NVP has multi-scale architecture, since it is not locally constrained, it fails to capture semantic representations at different levels.

\begin{figure}[htb]
\centering
\includegraphics[width=\linewidth]{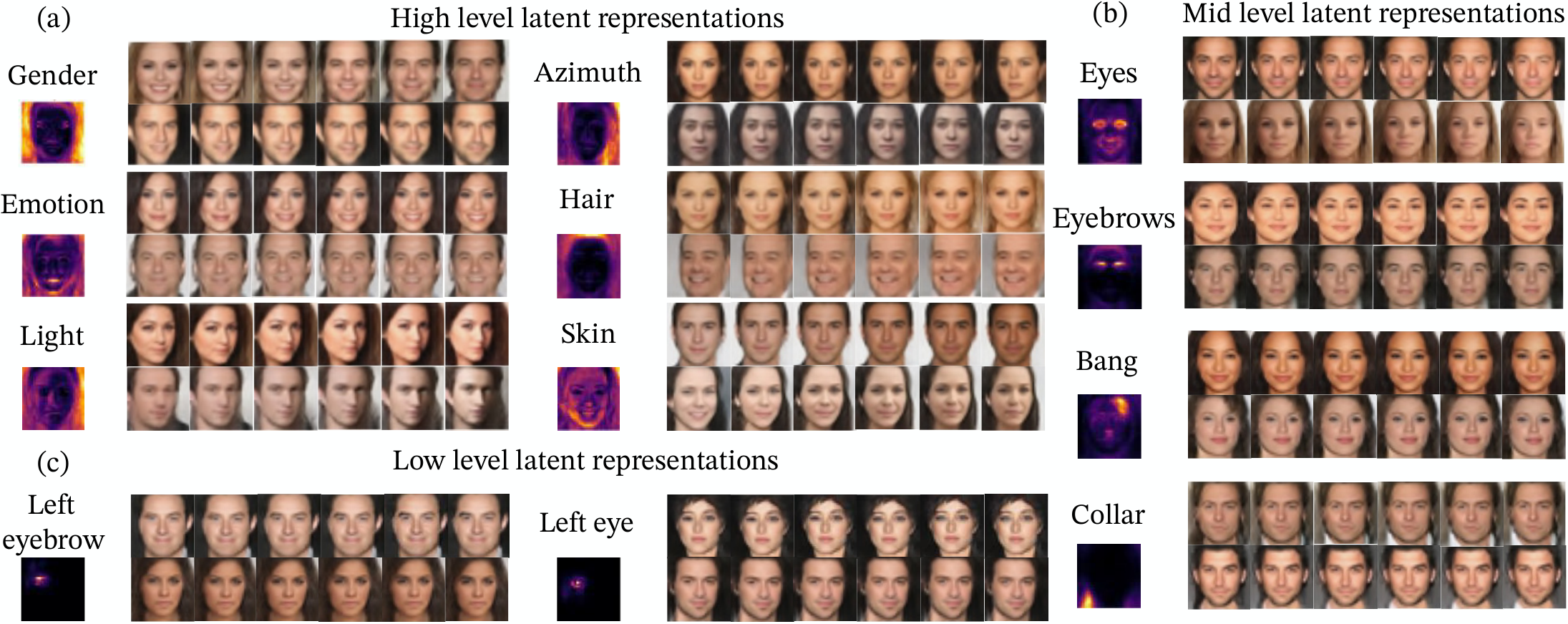}
\caption{
Semantic features of face images found at different levels.
(a) (Upper left) Six semantic features at the high level: gender, emotion, light, azimuth, hair, and skin. For each feature, the receptive field of the corresponding latent variable is plotted under its label, and to the right we plot the variation of two sample images when that latent variable is changed.
(b) (Right) Four semantic features at the middle level: eyes, eyebrows, bang, and collar.
(c) (Lower left) Two semantic features at the low level: left eyebrow and left eye.
}
\label{fig:factors}
\end{figure}

\subsection{Learned features on different scales} \label{sec:feature}

In this section, we show that some of those emergent structures correspond to explainable latent features. A flow-based generative model is a \emph{maximal encoding procedure}, because it uses bijective maps to preserve the number of variables before and after the encoding. It is believed that the images in the dataset live on a low-dimensional manifold, and we do not need all dimensions of the original dataset to encode them. In \fref{fig:RF}~(c) we show the histograms of the receptive fields' strengths, defined as $\text{RF}_l$ averaged over all pixels. Most latent variables produce receptive fields with small strengths, meaning that if we vary them the generated images will not be significantly affected. We focus on those latent variables with receptive field strengths greater than one, which have visible effects on the generated images. We use $h$ to label the RG level of latent variables. For example, the lowest-level latent variables have $h = 0$, whereas the highest-level ones have $h = 4$. Therefore, we focus on $h = 1$ (low-level), $h = 2$ (mid-level), and $h = 3$ (high-level).

For high-level latent representations, we found in total $30$ latent variables that have visible effects, and six of them are identified with disentangled and explainable semantics. Those factors are gender, emotion, light angle, azimuth, hair color, and skin color. In \fref{fig:factors}~(a), we plot the effects of varying those six high-level variables, together with their receptive fields. The samples are generated using mixed temperatures, as described in~\ref{append:temperature}. For mid-level latent representations, we plot four of them in \fref{fig:factors}~(b), which control eyes, eyebrows, hair bang, and collar respectively. For low-level representations, we plot two of them in \fref{fig:factors}~(c), controlling an eyebrow and an eye respectively. They are fully disentangled because their receptive fields do not overlap.

\begin{figure}[htb]
\centering
\includegraphics[width=\linewidth]{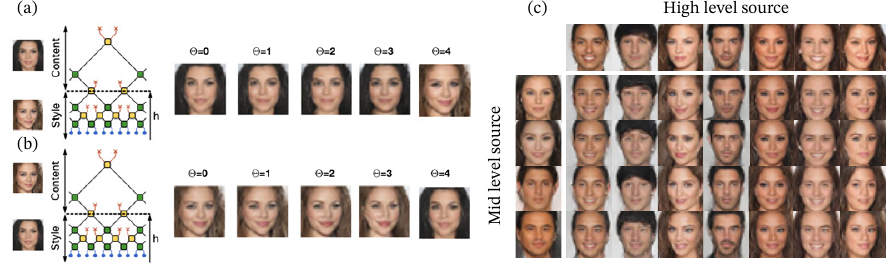}
\caption{
Image mixing in the hyperbolic tree-like latent space.
(a) The mixing between the high-level features of a brunette woman image and the low-level features of a blonde woman image. The two images are first mapped to hierarchical latent representations, then the latent variables with level $h < \Theta$ of the first image is replaced by those of the second image. Finally, the mixed latent variables are mapped back to a new image. Five mixed images with different threshold $\Theta$ are shown on the right.
(b) The same mixing procedure with the two images exchanged, where the blonde woman image provides high-level features and the brunette woman image provides low-level ones.
(c) More examples of image mixing. The seven images in the topmost row provide high-level representations, while the four in the leftmost column provide mid- and low-level ones. The images in the grid are mixed from them accordingly.
}
\label{fig:style_mix}
\end{figure}

\subsection{Image mixing in scaling direction}

Given two images $\vx_A$ and $\vx_B$, the conventional way of image mixing in the latent space takes a linear combination $\vz = \lambda \vz_A + (1 - \lambda) \vz_B$ between $\vz_A = G^{-1}(\vx_A)$ and $\vz_B = G^{-1}(\vx_B)$, with $\lambda \in [0, 1]$, and generates the mixed image $\vx = G(\vz)$. In RG-Flow, the direct access of the latent variables $\vz^{(h)}$ at any level $h$ of the hyperbolic tree-like latent space enables us to mix the images in a different manner, which we call ``hyperbolic mixing''.

We mix the large-scale (high-level) features of $\vx_A$ and the small-scale (low-level) features of $\vx_B$ by combining their latent variables at corresponding levels:
\begin{equation}
\vz^{(h)} = \left\{ \begin{array}{r@{\quad}cl}
\vz_A^{(h)} & \text{for} & h \geq \Theta, \\
\vz_B^{(h)} & \text{for} & h < \Theta,
\end{array} \right.
\label{eq:extrapolation}
\end{equation}
where $\Theta$ serves as a threshold of the scales. As shown in \fref{fig:style_mix}~(a), when we vary $\Theta$ from $0$ to $4$, more low-level information in the blonde-hair image is mixed with the high-level information in the black-hair image. Especially when $\Theta = 3$, the mixed face has similar eyebrows, eyes, nose, and mouth to the blonde-hair image, while the highest-level information, such as face orientation and hair color, is taken from the black-hair image. Note that this mixing is not symmetric under the interchange of $\vz_A$ and $\vz_B$, see \fref{fig:style_mix}~(b) for comparison. This hyperbolic mixing achieves a similar effect to StyleGAN~\cite{DBLP:conf/cvpr/KarrasLA19, DBLP:conf/cvpr/KarrasLAHLA20}, where we can mix the style information from one image and the content information from another. In \fref{fig:style_mix}~(c), we show more examples of mixing faces.

\begin{figure}[htb]
\centering
\includegraphics[width=\linewidth]{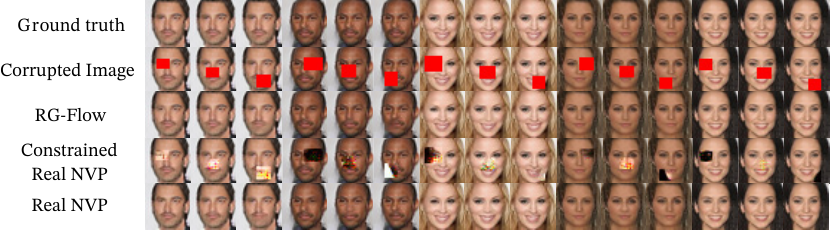}
\caption{Inpainting of locally corrupted images. The first row shows the ground truth images. The second row shows the locally corrupted images. The third row shows the images inpainted by RG-Flow. The fourth row shows the results by Real NVP when the same number of latent variables are allowed to be optimized. The last row shows the inpainting by Real NVP with all latent variables optimized.}
\label{fig:inpaint}
\end{figure}

\subsection{Image inpainting and error correction}

When we have a small and local region to be inpainted in the input image, the inference causal cone of RG-Flow ensures that at most $O(\log L)$ latent variables will be affected by that region. In \fref{fig:inpaint}, we show that RG-Flow can faithfully recover the corrupted region (marked as red) only using latent variables locating inside the inference causal cone, which are around one third of all latent variables. For comparison, if we randomly pick the same number of latent variables and put the constraint that we can only modify them in Real NVP, it fails to recover the region, as shown in \fref{fig:inpaint} (Constrained Real NVP). To achieve the recovery using Real NVP with a similar quality to RG-Flow, as shown in \fref{fig:inpaint} (Real NVP), all latent variables need to be modified, which are of $O(L^2)$ order. See~\ref{append:inpaint} for more details about the inpainting task and its quantitative evaluations.

\section{Discussion and conclusion}

In this paper, we have combined the ideas of renormalization group and sparse prior distribution to design RG-Flow, a flow-based generative model with hierarchical architecture and explainable latent space. In the analysis of its structure, we have shown that it can separate information in the input image at different scales and encode them in latent variables living on a hyperbolic tree. To visualize the latent representations in RG-Flow, we have defined the receptive fields for flow-based models in analogy to that for CNN. The experiments on synthetic datasets have shown the disentanglement of representations at different levels and spatial positions, and the receptive fields serve as a visual guidance to find explainable representations. Richer applications have been demonstrated on the CelebA dataset, including efficient image inpainting that utilizes the locality of the image, and mixing of style and content. In contrast, such semantically meaningful representations of mid-level and low-level structures do not emerge in globally connected flow models like Real NVP. The versatile architecture of RG-Flow can be incorporated with any bijective map to further improve its expressive power.

In RG-Flow, two low-level representations are fully disentangled if their receptive fields do not overlap. For high-level representations with large receptive fields, we use a sparse prior to encourage their disentanglement. However, we find that if the dataset only contains few high-level factors, such as the 3D Chairs dataset~\cite{DBLP:conf/cvpr/AubryMERS14} shown in~\ref{append:other_datasets}, it is hard to find disentangled and explainable high-level representations, because of the redundant nature of the maximal encoding in flow-based models. Incorporating information theoretic criteria to disentangle high-level representations will be an interesting future direction.

\section{Acknowledgements}

H~Y~H and Y~Z~Y are supported by a UC Hellman Fellowship. H~Y~H is also grateful for the supporting from Swarma-Kaifeng Project which is sponsored by Swarma Club and Kaifeng Foundation. Y~C and B~O are supported by NSF-IIS-1718991.

\section{Data availability statement}

Our code is available at \url{https://github.com/hongyehu/RG-Flow} \\
The data that support the findings of this study are openly available at the following URL/DOI: \\
CelebA: \url{https://mmlab.ie.cuhk.edu.hk/projects/CelebA.html} \\
CIFAR-10: \url{https://www.cs.toronto.edu/kriz/cifar.html} \\
3D Chairs: \url{https://www.di.ens.fr/willow/research/seeing3Dchairs/} \\

\appendix

\section{Details of network and training procedure} \label{append:net}

\begin{figure}[htb]
\centering
\includegraphics[width=0.9\linewidth]{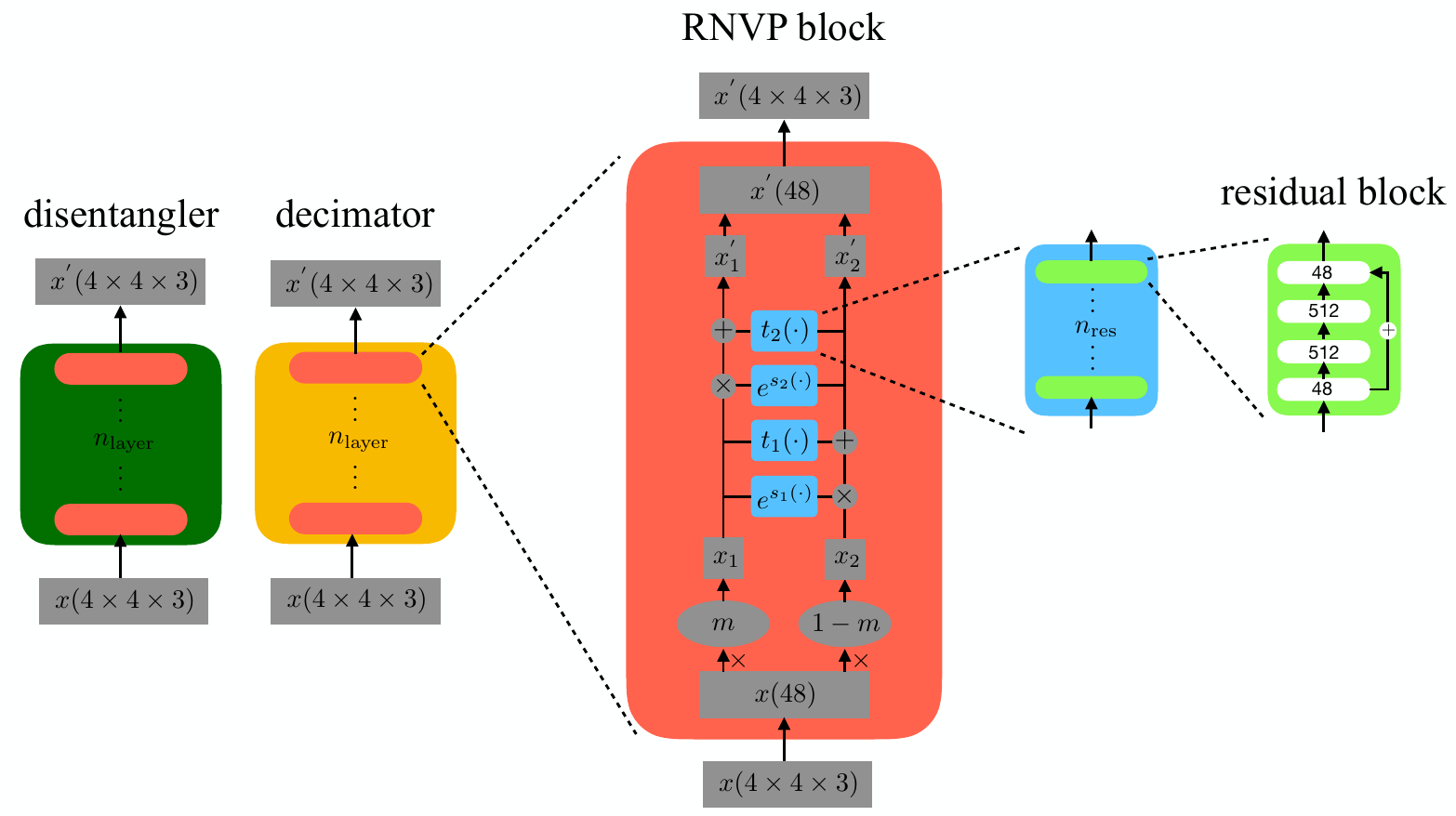}
\caption{The details of a disentangler or decimator in our network.}
\label{fig:network_detail}
\end{figure}

As shown in \fref{fig:structure}~(a), each RG step includes a disentangler (green block) and a decimator (yellow block), which is parameterized by a bijective neural network. The blocks in every horizontal level share parameters, so we view them as a same block in the discussions. In our reported experiments, disentanglers and decimators in different RG steps do not share parameters. However, we can also make them share parameters, which implements a scale-invariant course-graining process.

For each disentangler or decimator, we spilt the image into $4 \times 4$ patches as shown in \fref{fig:structure}~(b), stack them along the batch dimension, feed them into the network, and merge the output patches into a new image. After each RG step, the edge length of the image is halved (the number of black dots in a row, compared to red crosses), except for the last (highest-level) RG step that decimates all variables.

The choice of the bijective neural networks for the disentanglers and decimators can be versatile, and the performance of RG-Flow strongly depends on them. Since tuning the expressive power of those blocks is not the focus of our work, we take the coupling layer from Real NVP~\cite{DBLP:conf/iclr/DinhSB17}, denoted as RNVP block in the following, which has great expressive power and is easy to invert.

\Fref{fig:network_detail} illustrates the architecture of a disentangler or decimator in our implementation. Each RNVP block is shown as a red block. It takes a $4 \times 4$ image patch $x$ as input, and split it into $x_1$ and $x_2$ using the checkerboard mask $m$. They are coupled to produce the output $x'_2$, using the formula
\begin{equation}
x'_2 = x_2 \odot \exp \left( s_1(x_1) \right) + t_1(x_1),
\end{equation}
where $\odot$ is the element-wise product. The scale network $s(\cdot)$ and the translation network $t(\cdot)$ can be arbitrarily complex to enhance the expressive power, as long as the number of variables is the same for their input and output. Then we use $x'_2$ to alter $x_1$ in the similar manner and outputs $x'_1$, and combine them to produce the output $x'$ of the RNVP block.

We use residual networks with $n_\text{res}$ residual blocks as $s(\cdot)$ and $t(\cdot)$, shown as blue blocks in \fref{fig:network_detail}, and we choose $n_\text{res} = 4$ in our implementation. Each residual block has $3$ linear layers with size $24 \to 512, 512 \to 512, 512 \to 24$. where $24 = \frac{1}{2} \times 4 \times 4 \times 3$ is the number of variables in a masked image patch. Between linear layers, we insert SiLU activations~\cite{DBLP:journals/nn/ElfwingUD18}, which is reported to give better results than ReLU and softplus, and its smoothness benefits our further analysis with higher order derivatives. We use Kaiming initialization~\cite{DBLP:conf/cvpr/HeZRS16} and weight normalization~\cite{DBLP:conf/nips/SalimansK16} on the linear layers.

The CelebA dataset contains rich information at different scales, including high-level information like gender and emotion, mid-level one like shapes of eyes and nose, and low-level one like details in hair and wrinkles. Because lower-level RG steps take larger images as input, we heuristically put more parameters in them. The numbers of RNVP blocks in all RG steps are $n_\text{layer} = 8, 6, 4, 2$ respectively.

To preprocess the dataset, we use the aligned images from CelebA, crop a $148 \times 148$ patch at the center of each image, downscale the patch to $32 \times 32$ using bicubic downsampling, and randomly flip it horizontally.

We use AdamW optimizer~\cite{DBLP:conf/iclr/LoshchilovH19} with conventional learning rate $10^{-3}$ and weight decay rate $5 \times 10^{-5}$. To further stabilize training, we use gradient clipping with global norm $1$. Between coupling layers, we use checkpointing~\cite{DBLP:conf/nips/PaszkeGMLBCKLGA19} to reduce memory usage. The maximal batch size can be set to $1,024$ on an Nvidia Titan RTX given the current setup, which approximately has a million parameters. In our experiment, the batch size is conventionally set to $64$, and a training step takes about $1.2$ seconds.

\section{Details of synthetic multi-scale datasets} \label{append:MSDS}

\begin{figure}[htb]
\centering
\includegraphics[width=0.9\linewidth]{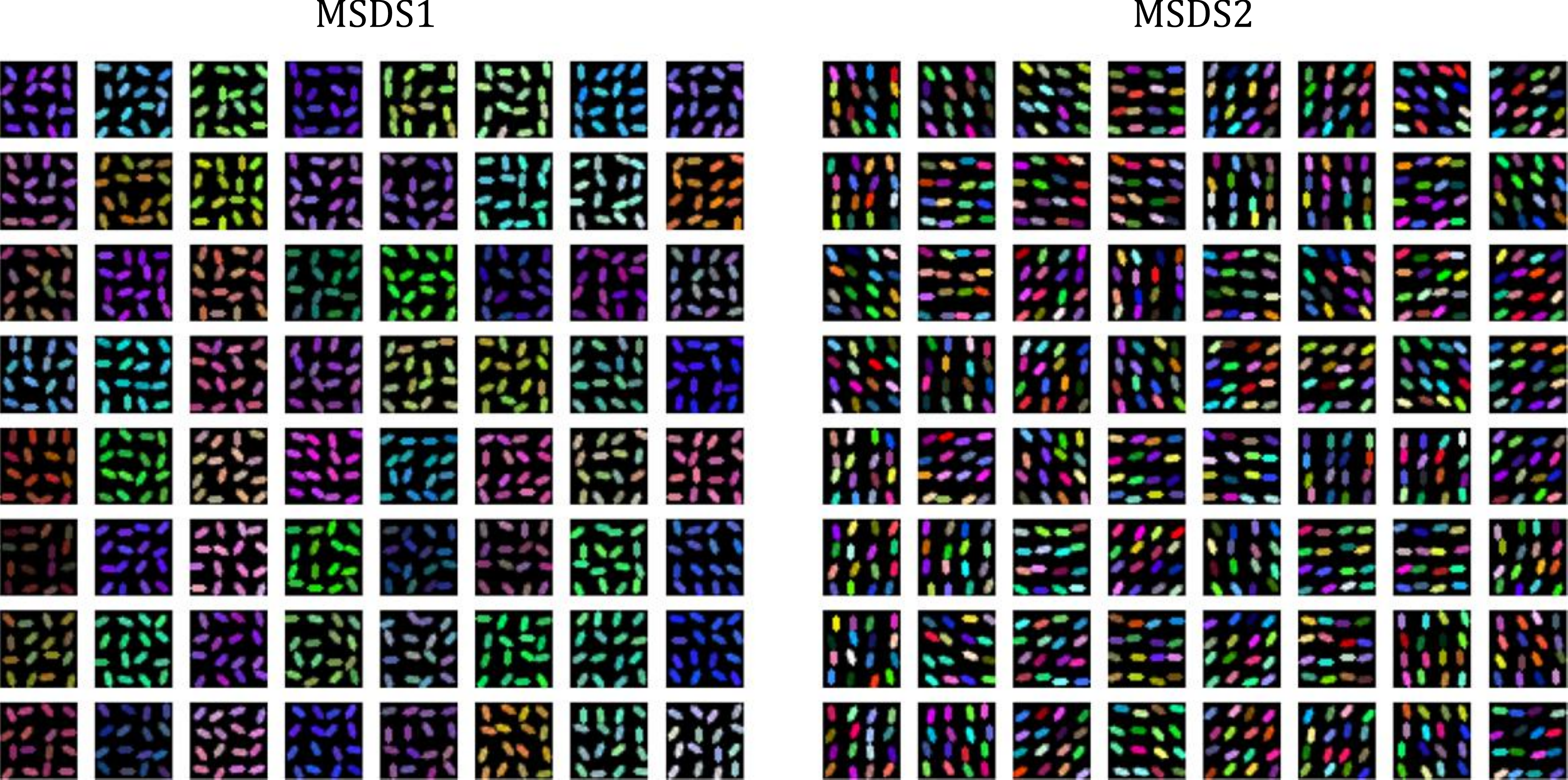}
\caption{Samples from MSDS1 and MSDS2.}
\label{fig:msds_ground_truth}
\end{figure}

To illustrate RG-Flow's ability to disentangle representations at different scale and spatial positions, we propose two synthetic image datasets with multi-scale features, which we name MSDS1 and MSDS2, as shown in \fref{fig:msds_ground_truth}. Each dataset contains $10^5$ images of $32 \times 32$ pixels. In each image, there are $16$ ovals with different colors and orientations, and their positions have small random variations to deform the $4 \times 4$ grid. In MSDS1, all ovals in a same image have almost the same color, while their orientations are randomly distributed, so the color is a global feature and the orientations are local ones. In MSDS2, on the contrary, the orientation is a global feature and the colors are local ones.

\begin{figure}[htb]
\centering
\includegraphics[width=\linewidth]{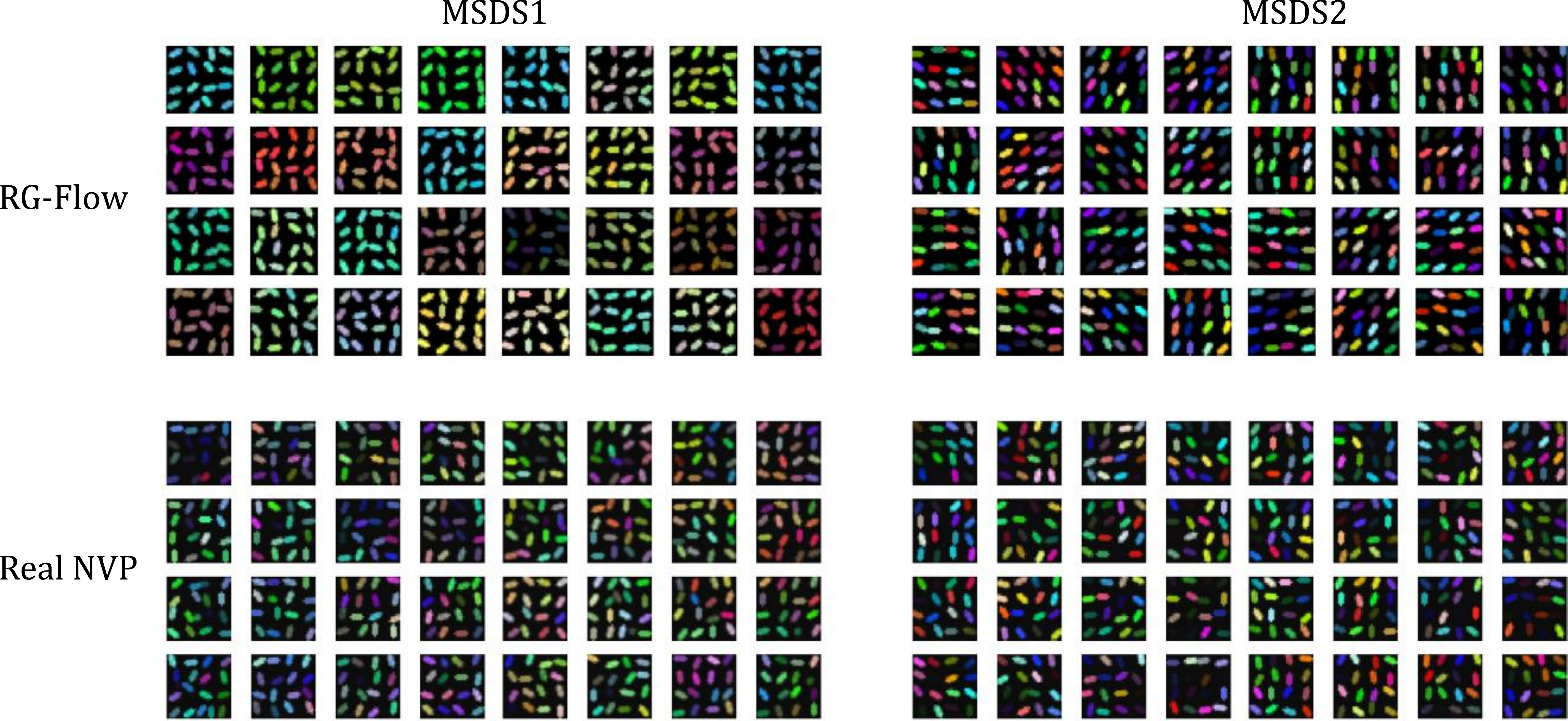}
\caption{Samples from RG-Flow and Real NVP trained on MSDS1 and MSDS2.}
\label{fig:msds_generated}
\end{figure}

We train RG-Flow on the two datasets, with $n_\text{layer} = 4$ and $n_\text{res} = 4$, and other hyperparameters described in~\ref{append:net}. For comparison, we also train Real NVP on those datasets, with approximately the same number of trainable parameters. Their generated images are shown in \fref{fig:msds_generated}, where we can intuitively see that RG-Flow has learned the characteristics of the two datasets. Namely, the ovals in each image from MSDS1 have almost the same color, and from MSDS2 the same orientation. In contrast, Real NVP fails to capture the global and the local features of those datasets. The metrics of bits per dimension (BPD) and Fréchet Inception distance (FID) are listed in \tref{table:msds_bpd}. Note that FID may not reflect much semantic property for such synthetic datasets. The results also show that RG-Flow with Laplacian prior captures the disentangled representations better than that with Gaussian prior because of the sparsity of Laplacian distribution, which is discussed more in~\ref{append:sparse_prior}.

\begin{table}[htb]
\begin{center}
\begin{tabular}{lcccc}
\toprule
& \multicolumn{2}{c}{BPD $\downarrow$} & \multicolumn{2}{c}{FID $\downarrow$} \\
& MSDS1 & MSDS2 & MSDS1 & MSDS2 \\
\midrule
RG-Flow + Laplacian prior & \textbf{0.906} & \textbf{1.01} & \textbf{1.16} & \textbf{2.61} \\
RG-Flow + Gaussian prior & 1.60 & 1.53 & 7.42 & 7.16 \\
Real NVP & 1.07 & 1.14 & 50.4 & 78.0 \\
\bottomrule
\end{tabular}
\caption{BPD and FID from RG-Flow and Real NVP trained on MSDS1 and MSDS2.}
\label{table:msds_bpd}
\end{center}
\end{table}

\section{Receptive fields of Real NVP} \label{append:RNVP}

\begin{figure}[htb]
\centering
\includegraphics[width=\linewidth]{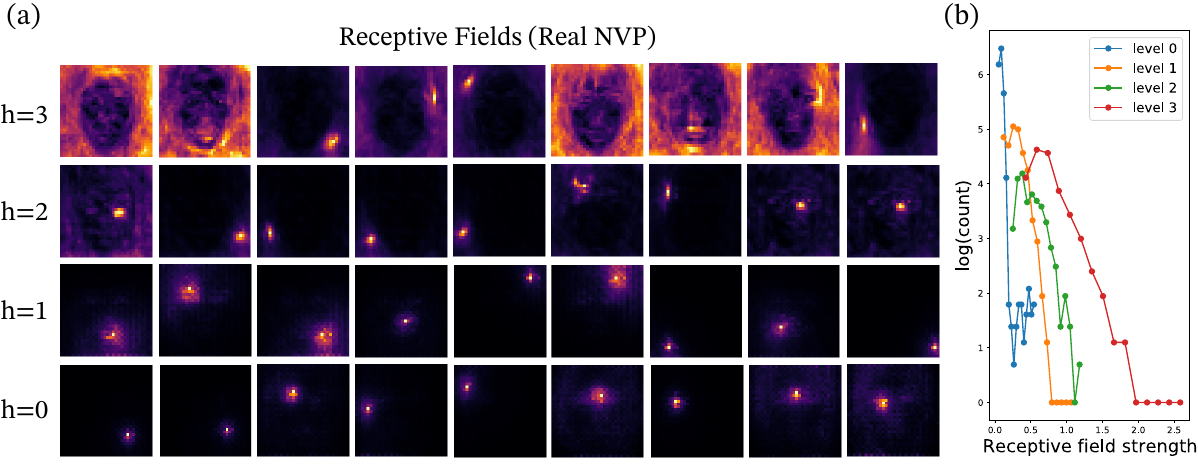}
\caption{
(a) Some randomly picked receptive fields from Real NVP trained on CelebA at different levels.
(b) Histograms of the receptive field strengths at each level.
}
\label{fig:RF_RNVP}
\end{figure}

As a comparison to \fref{fig:RF}, we plot receptive fields of Real NVP in \fref{fig:RF_RNVP}~(a), together with the histograms of their strengths in \fref{fig:RF_RNVP}~(b). Without the constraint of locality on the bijective maps, there is no generation causal cone for Real NVP and other globally connected multi-scale models, and we cannot find semantically meaningful representations emerging at low levels. In contrast, RG-Flow can seperate semantic information at different scales, as shown in \fref{fig:RF}.

\section{Image generation using effective and mixed temperatures} \label{append:temperature}

As a common practice in generative models, we can define the effective temperature for RG-Flow. Our prior distribution is
\begin{equation}
p_Z^{*}(\vz) = \prod_l p(z_l),
\end{equation}
where $l = (h, i, j, c)$ labels every latent variable by its RG level $h$, spatial position $(i, j)$, and color channel $c$. For a model with effective temperature $T$ ($T > 0$), the prior distribution is changed to
\begin{equation}
p_{Z, T}^{*}(\vz) \propto \left( p_Z^{*}(\vz) \right)^\frac{1}{T}.
\end{equation}
For Laplacian prior, the effective temperature is implemented as
\begin{equation}
p_{Z, T}^{*}(\vz) = \prod_l \frac{1}{2 T} \exp \left( -\frac{|z_l|}{T} \right).
\end{equation}
Moreover, thanks to the factorized form of the prior distribution, we can define a mixed temperature model by
\begin{equation}
p_{Z, \{T_l\}}^{*}(\vz) \propto \prod_l \left( p(z_l) \right)^\frac{1}{T_l},
\end{equation}
where $T_l$ can be different for each latent variable.

In our training procedure, we find the bijective maps at the lowest level take a longer time to converge. Therefore, when plotting the results in \fref{fig:factors}, we use the mixed temperature scheme with $T_{h = 0} = 0.2$, $T_{h = 1} = T_{h = 2} = T_{h = 3} = 0.6$, where $T_h$ is the effective temperature at level $h$. Then we vary each latent variable from $0$ to $6$ if $h = 1$, and from $0$ to $1.5$ if $h = 2$ or $3$.

\section{Importance of sparse prior distribution} \label{append:sparse_prior}

\begin{figure}[htb]
\centering
\includegraphics[width=\linewidth]{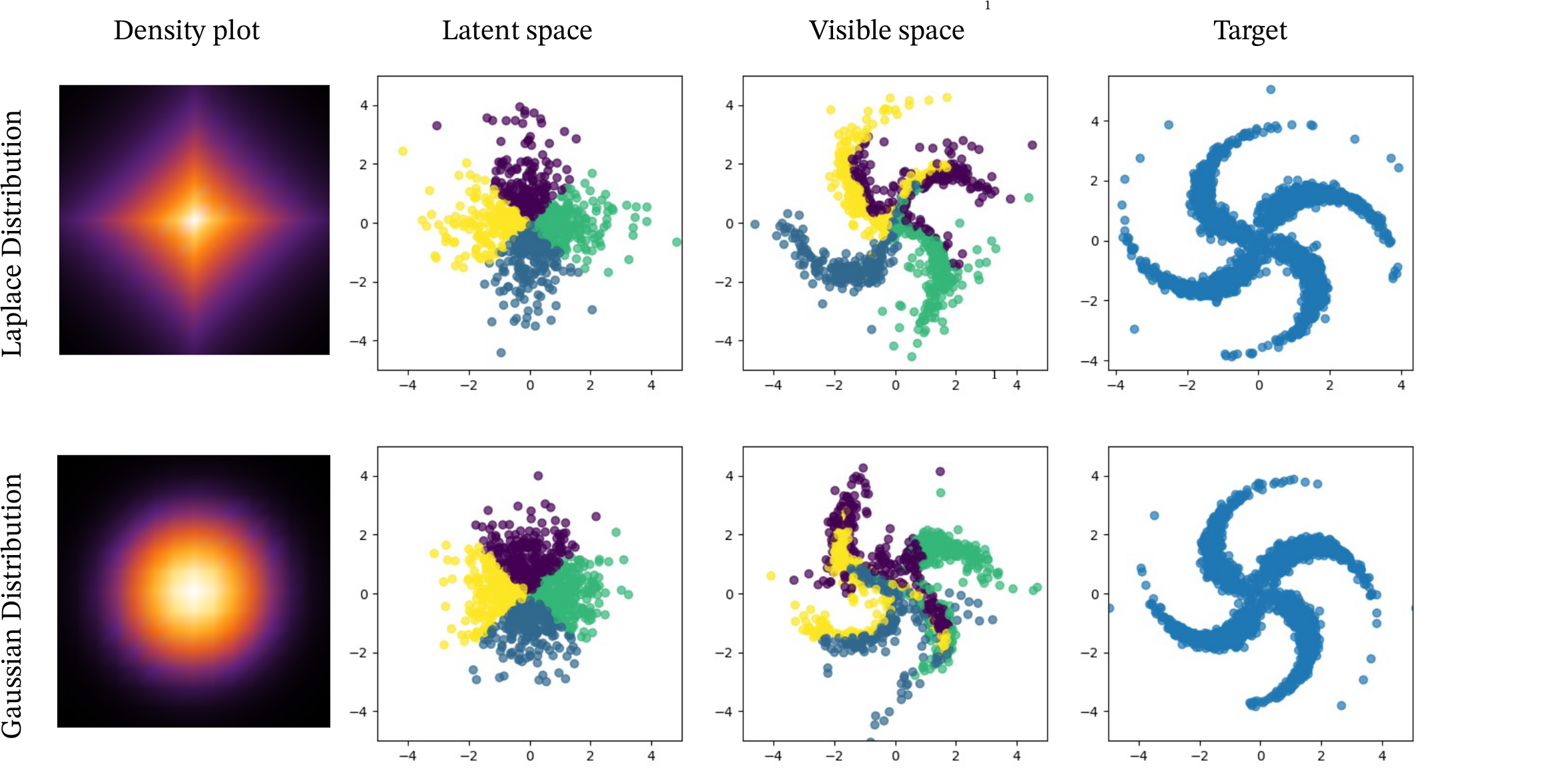}
\caption{Two-dimensional pinwheel distribution.}
\label{fig:toy_model}
\end{figure}

In the first column of \fref{fig:toy_model}, the density plots of Laplacian distribution and Gaussian distribution are shown. As we can see, the Gaussian distribution has rotational symmetry, whereas the Laplacian distribution breaks $SO(2)$ rotational symmetry to $C_4$ symmetry. A flow-based generative model is trained to match the target distribution, which is a four-leg pinwheel as shown in the last column in \fref{fig:toy_model}. Given either Gaussian distribution or Laplacian distribution as the prior distribution, the model can learn the target distribution. In the second column, we sample $100$ points and color them by their living quadrant in the prior distribution. Then we map them to the observable space using the trained model, as shown in the third column. We see that the four quadrants of the latent space are approximately mapped to the four legs of the pinwheel in the observable space if Laplacian prior is used. In the case of Gaussian prior, since it has rotational symmetry, the points in different quadrants are mixed more in the observable space, which makes it harder to explain the mapping. The quantitative comparison of the performance between Laplacian and Gaussian distributions on the MSDS datasets are reported in \tref{table:msds_bpd}.

\section{Details of inpainting experiments} \label{append:inpaint}

For the inpainting experiments shown in \fref{fig:inpaint}, we randomly choose a corrupted region of $10 \times 10$ pixels on the ground truth image, marked as the red patch in the second row of \fref{fig:inpaint}. We generate an image $\vx_g$ from latent variables $\vz_g$, and use its corresponding region to fill in the corrupted region. Then we map the filled image $\vx_f$ back to the latent variables $\vz_f$ and compute its log-likelihood. To recover the ground truth image, we optimize $\vz_g$ to maximize the log-likelihood.

For RG-Flow, we only vary the latent variables living inside the inference causal cone, which are about $1,200$ out of $3,072$ latent variables. For the constrained Real NVP, we randomly pick the same amount of latent variables and allow them to be optimized, and we find it fails to inpaint the image in general. As a check, we find Real NVP can successfully inpaint the images if we optimize all latent variables, as shown in the last row of \fref{fig:inpaint}.

We use the conventional Adam optimizer to do the optimization. During the optimization procedure, we find the optimizer can be trapped in local minima. Therefore, for all experiments, we first randomly draw $200$ initial samples of latent variables that are allowed to be optimized, then pick the one with the largest log-likelihood as the initialization.

To quantitatively evaluate the quality of inpainted images, we compute the peak signal-to-noise ratio (PSNR) of them against the ground truth images, and take the average over the $15$ samples shown in \fref{fig:inpaint}. To further incorporate semantic properties in the evaluation, we also use the Inception-v3 network~\cite{DBLP:conf/cvpr/SzegedyVISW16} to extract features from the images as in FID score, and compute the PSNR. The results are listed in \tref{table:inpaint}.

\begin{table}[htb]
\begin{center}
\begin{tabular}{lcc}
\toprule
& PSNR $\uparrow$ & Inception-PSNR $\uparrow$ \\
\midrule
RG-Flow & \textbf{38.7} & \textbf{24.9} \\
Constrained Real NVP & 23.2 & 15.7 \\
Real NVP & 37.0 & 24.8 \\
\bottomrule
\end{tabular}
\caption{PSNR and Inception-PSNR of inpainted images.}
\label{table:inpaint}
\end{center}
\end{table}

\section{Experiments on other datasets} \label{append:other_datasets}

\begin{figure}[p]
\centering
\includegraphics[width=\linewidth]{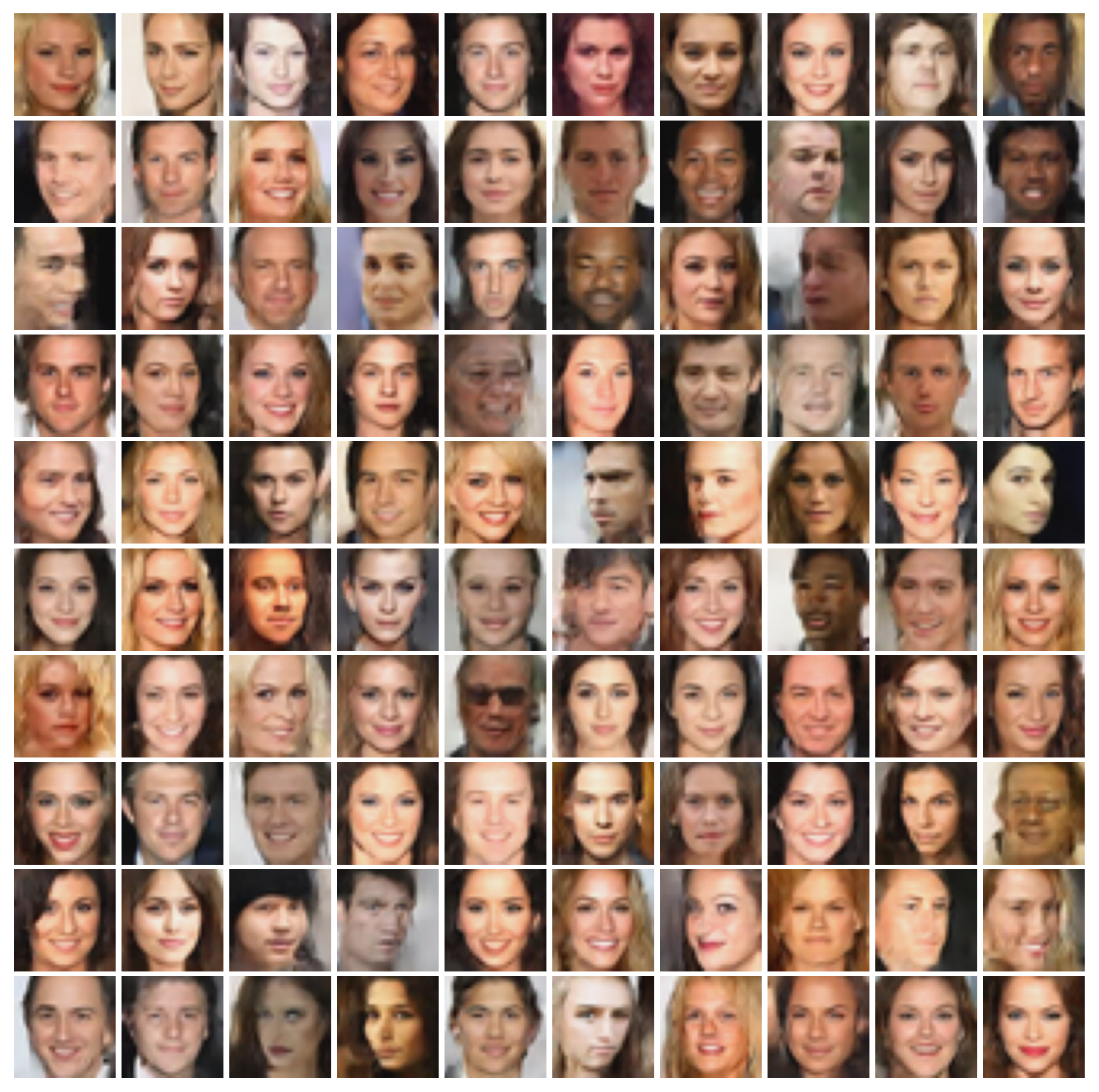}
\caption{Samples from RG-Flow trained on CelebA dataset. We use $T = 0.9$ when sampling.}
\label{fig:celeba_sample}
\end{figure}

\begin{figure}[p]
\centering
\includegraphics[width=\linewidth]{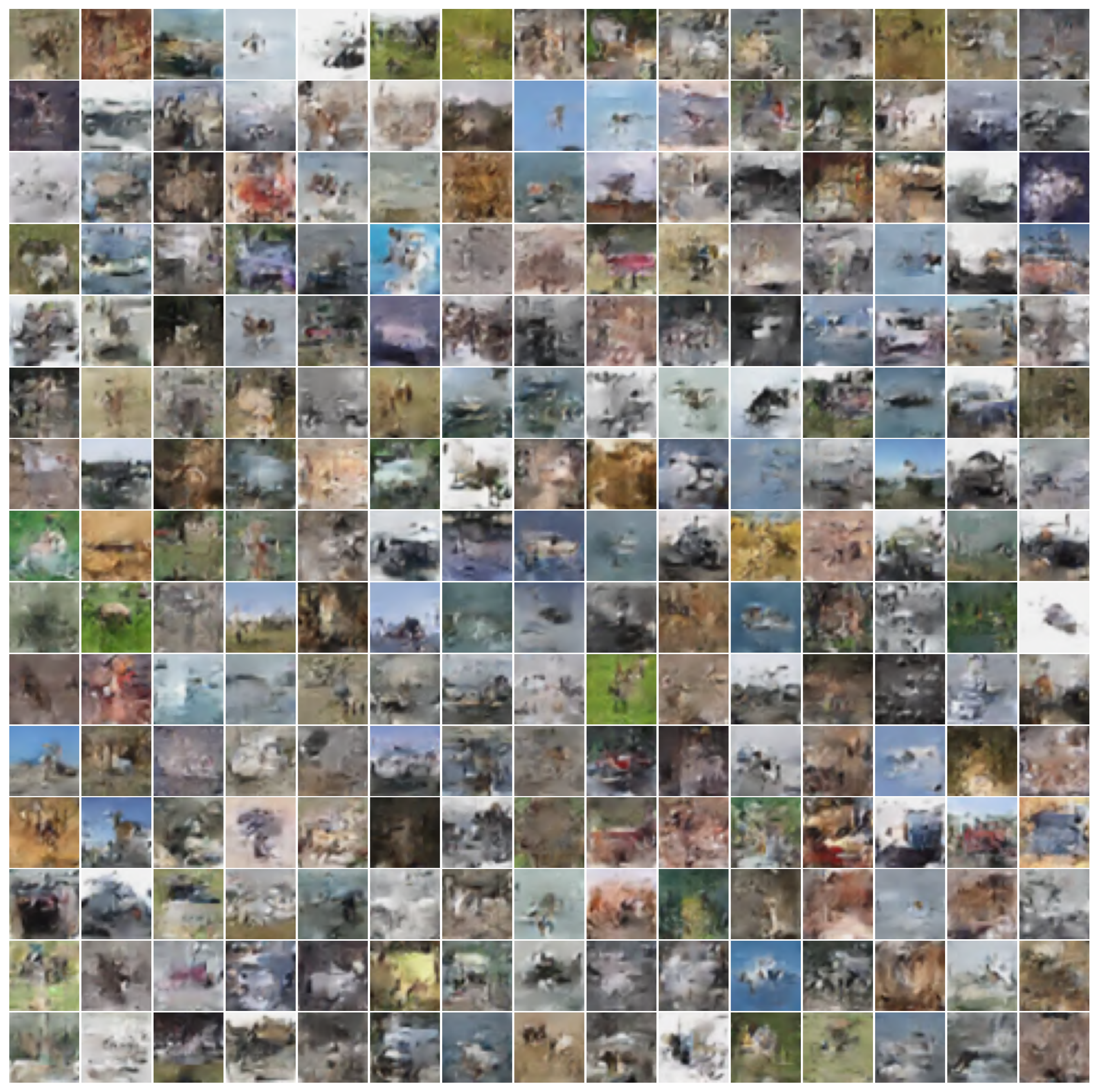}
\caption{Samples from RG-Flow trained on CIFAR-10 dataset.}
\label{fig:cifar10_sample}
\end{figure}

\begin{figure}[p]
\centering
\includegraphics[width=\linewidth]{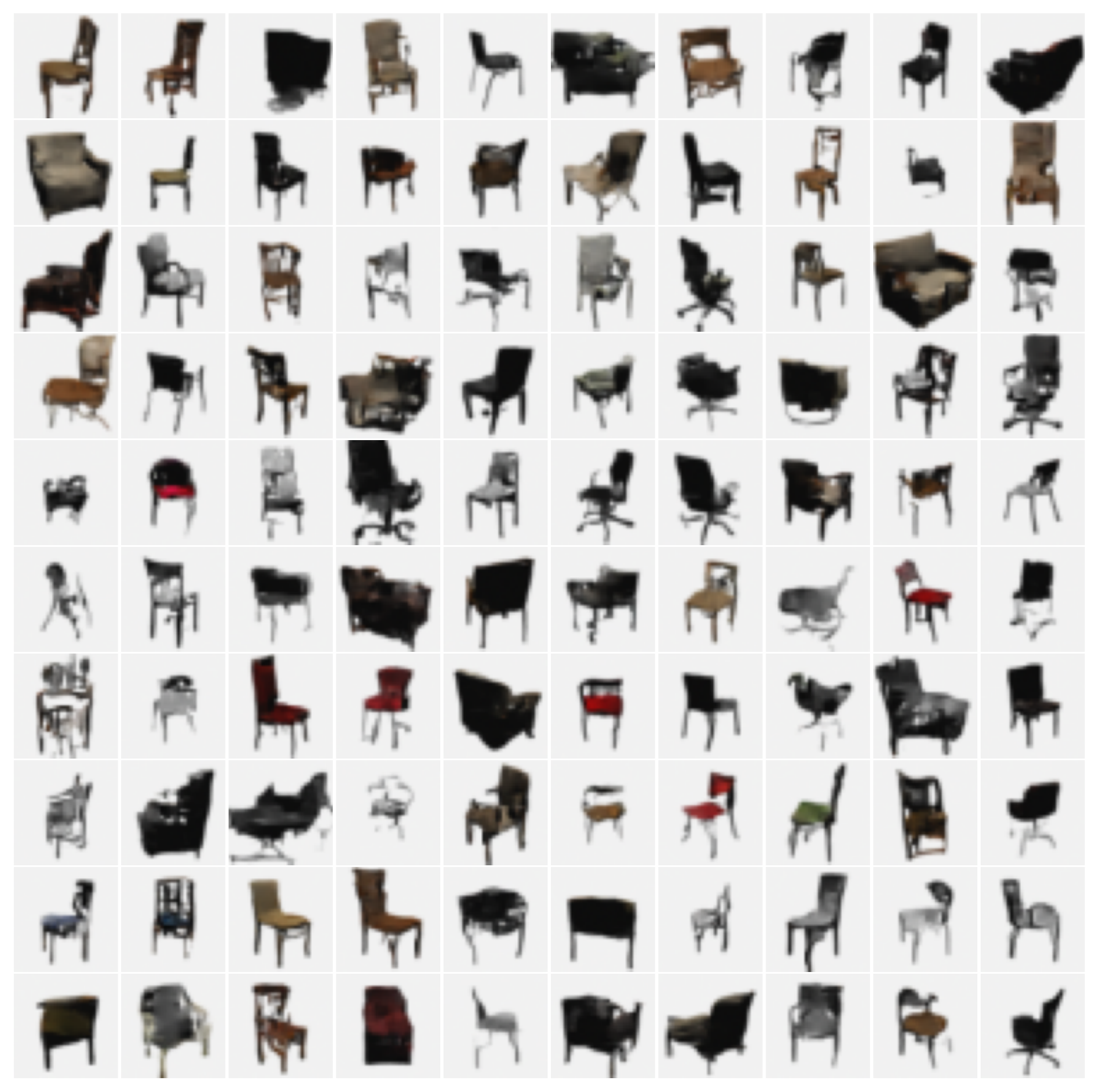}
\caption{Samples from RG-Flow trained on 3D Chairs dataset.}
\label{fig:chair_sample}
\end{figure}

For CIFAR-10 dataset~\cite{cifar10}, we use the same hyperparameters as described in~\ref{append:net}. For 3D Chairs dataset~\cite{DBLP:conf/cvpr/AubryMERS14}, we use $n_\text{layer} = 8$ for all RG steps, because there is not so much low-level information as in CelebA and CIFAR-10. We also train Real NVP on those datasets for comparison, with approximately the same number of trainable parameters. The metrics of bits per dimension (BPD) and Fréchet Inception distance (FID) are listed in \tref{table:bpd}. The samples from each dataset are shown in figures~\ref{fig:celeba_sample},~\ref{fig:cifar10_sample}, and~\ref{fig:chair_sample}.

\begin{table}[htb]
\begin{center}
\begin{tabular}{lcccccc}
\toprule
& \multicolumn{3}{c}{BPD $\downarrow$} & \multicolumn{3}{c}{FID $\downarrow$} \\
& CelebA & CIFAR-10 & 3D Chairs & CelebA & CIFAR-10 & 3D Chairs \\
\midrule
RG-Flow & 3.47 & \textbf{3.35} & \textbf{0.930} & 31.3 & \textbf{98.5} & \textbf{36.2} \\
Real NVP & \textbf{3.36} & 3.61 & 0.933 & \textbf{11.1} & 126 & 73.8 \\
\bottomrule
\end{tabular}
\caption{Bits per dimension (BPD) from RG-Flow and Real NVP trained on various datasets.}
\label{table:bpd}
\end{center}
\end{table}

\clearpage

\section{More receptive fields of latent representations} \label{append:RF}

\begin{figure}[htb]
\centering
\includegraphics[width=0.7\linewidth]{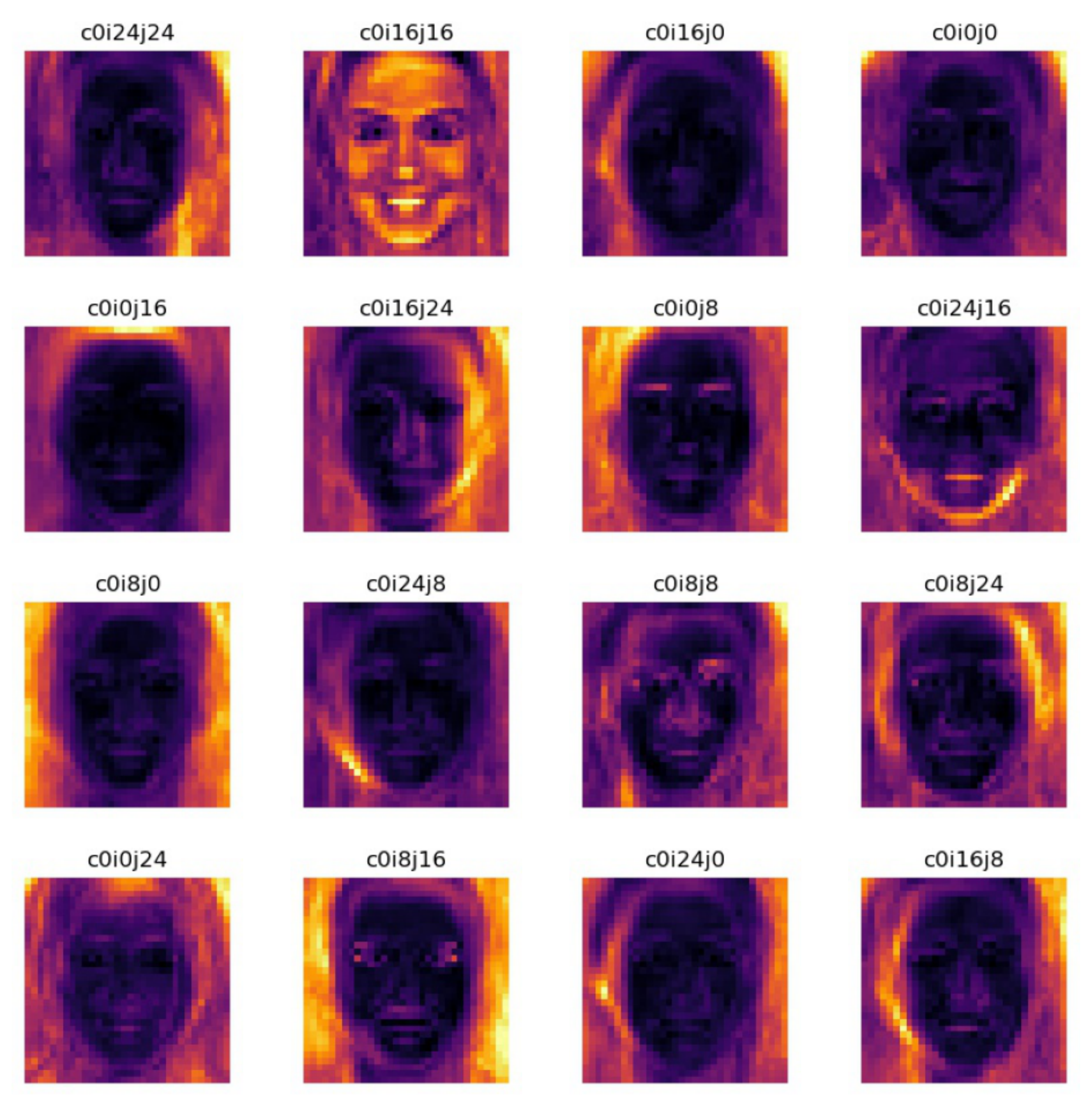}
\caption{Receptive fields of high-level latent variables ($h = 3$).}
\label{fig:rf_layer3}
\end{figure}

\begin{figure}[p]
\centering
\includegraphics[width=\linewidth]{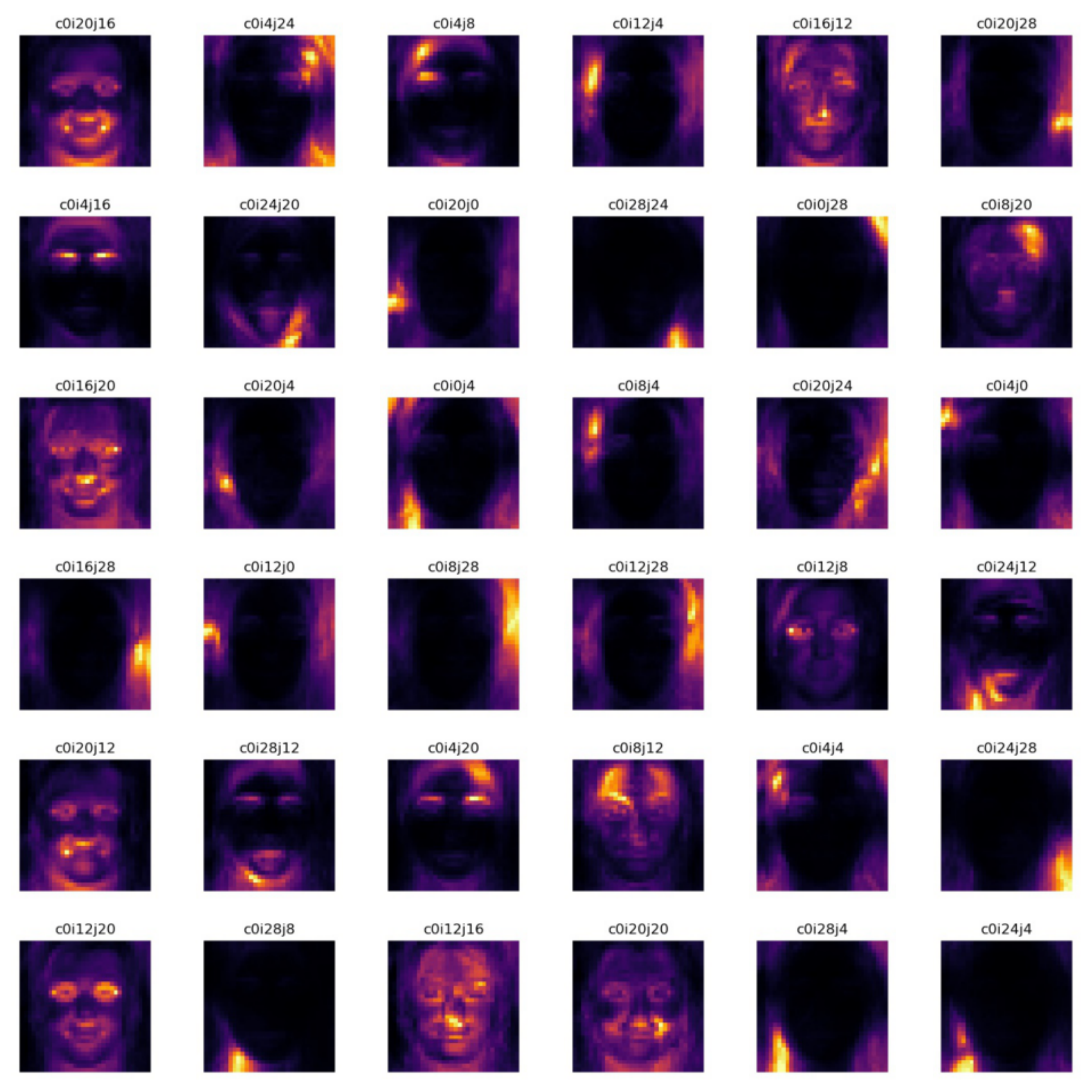}
\caption{Receptive fields of mid-level latent variables ($h = 2$).}
\label{fig:rf_layer2}
\end{figure}

\begin{figure}[p]
\centering
\includegraphics[width=\linewidth]{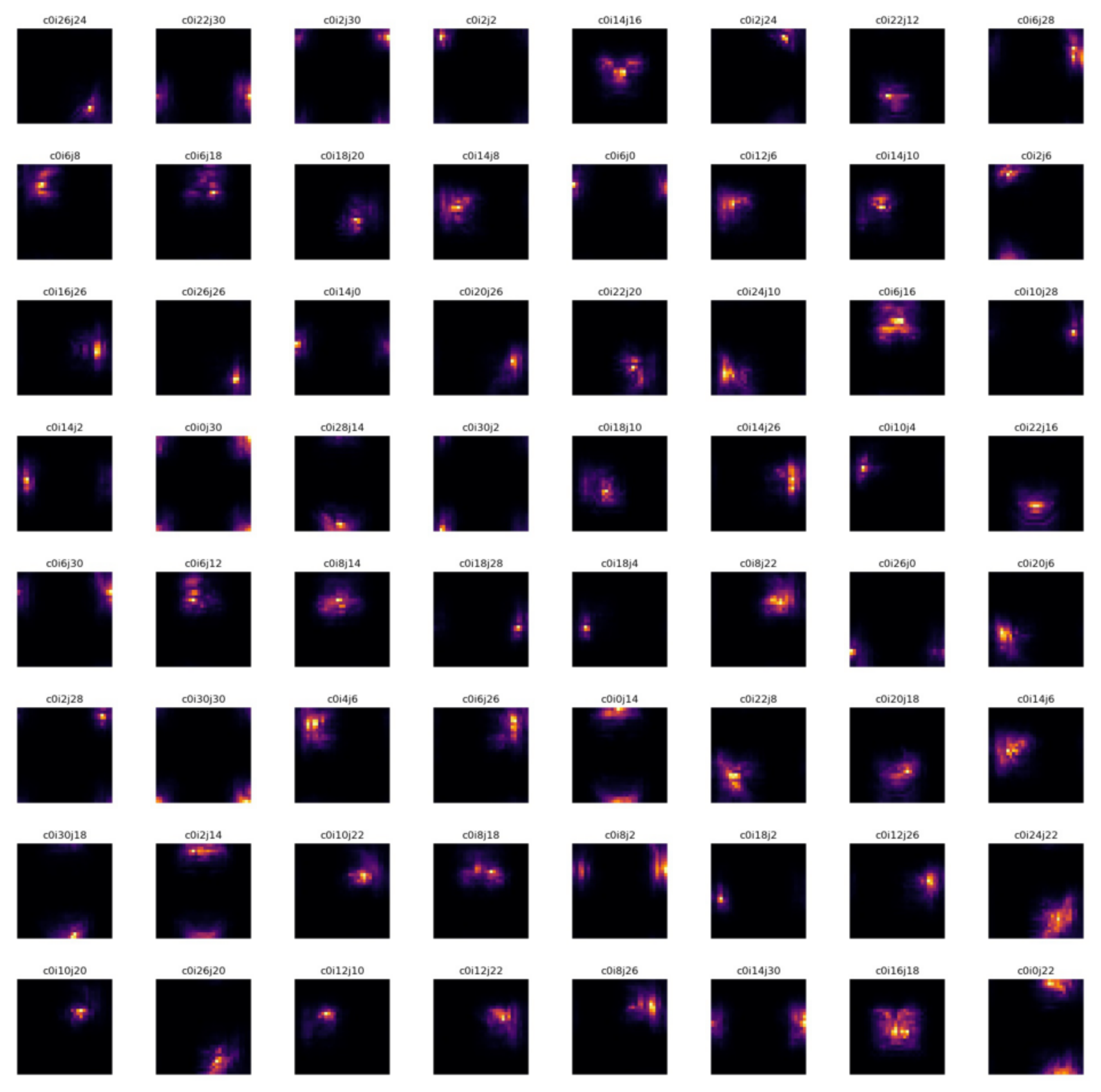}
\caption{Receptive fields of low-level latent variables ($h = 1$).}
\label{fig:rf_layer1}
\end{figure}

\begin{figure}[p]
\centering
\includegraphics[width=\linewidth]{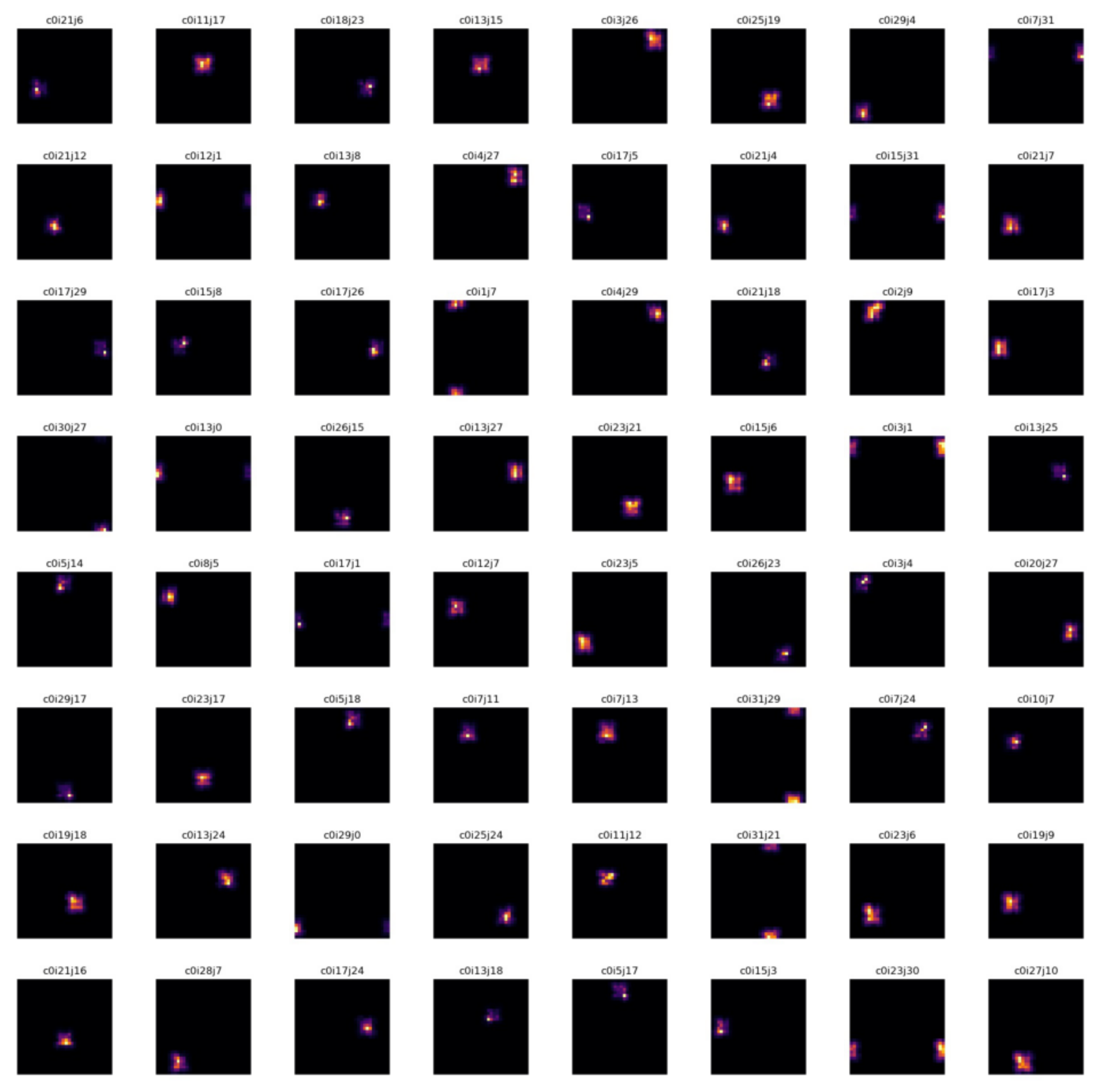}
\caption{Receptive fields of the lowest-level latent variables ($h = 0$).}
\label{fig:rf_layer0}
\end{figure}

More examples of randomly chosen receptive fields are plotted in figures~\ref{fig:rf_layer3},~\ref{fig:rf_layer2},~\ref{fig:rf_layer1}, and~\ref{fig:rf_layer0}. For better visualization, we normalize each receptive field's strength to one.

\clearpage


\providecommand{\newblock}{}

\end{document}